\documentclass{article}

\PassOptionsToPackage{numbers}{natbib}
\usepackage[main, final]{neurips_2026}

\usepackage[utf8]{inputenc} % allow utf-8 input
\pdfmapline{+ectt1000 SFTT1000 "T1Encoding ReEncodeFont" <cm-super-t1.enc <sftt1000.pfb}

\usepackage[T1]{fontenc}    % use 8-bit T1 fonts
\usepackage{hyperref}       % hyperlinks
\usepackage{url}            % simple URL typesetting
\usepackage{booktabs}       % professional-quality tables
\usepackage{amsfonts}       % blackboard math symbols
\usepackage{nicefrac}       % compact symbols for 1/2, etc.
\usepackage{microtype}      % microtypography
\usepackage{xcolor}         % colors
\usepackage{amsmath}

\usepackage{wrapfig}
\usepackage{subcaption}
\usepackage{multirow}
\usepackage{graphicx}
\usepackage{siunitx}
\usepackage{booktabs}
\usepackage{svg}
\usepackage{xcolor}
\usepackage{caption}
\definecolor{firstcolor}{RGB}{220,20,60}    % red (1st place)
\definecolor{secondcolor}{RGB}{30,144,255}  % blue (2nd place)
\definecolor{thirdcolor}{RGB}{34,139,34}    % green (3rd place)
\newcommand{\first}[1]{\textcolor{firstcolor}{\textbf{#1}}}
\newcommand{\second}[1]{\textcolor{secondcolor}{\textbf{#1}}}
\newcommand{\third}[1]{\textcolor{thirdcolor}{\textbf{#1}}}

\title{ControlFlow3D: Distilling Multi-View Knowledge into Latent Flow Matching for Point Cloud Upsampling}

\author{%
  \bfseries Yuang Liu\textsuperscript{1}\thanks{These authors contributed equally. The co-first authors are listed in alphabetical order by surname.} \quad
  Zhi Zuo\textsuperscript{1,3,\ensuremath{\ast}}\thanks{Project lead.} \quad
  Zhengkai Zhao\textsuperscript{1} \quad
  Lirui Zhang\textsuperscript{1} \\[0.35em]
  \bfseries Pan Gao\textsuperscript{1}\thanks{Corresponding authors.} \quad
  Hao Feng\textsuperscript{2} \quad
  Zhengzhe Liu\textsuperscript{2\ensuremath{\ddagger}} \\[1.1em]
  {\normalfont\normalsize
   \textsuperscript{1}Nanjing University of Aeronautics and Astronautics} \\[0.25em]
  {\normalfont\normalsize
   \textsuperscript{2}Lingnan University \qquad
   \textsuperscript{3}Weidian Corporation} \\[0.7em]
  {\normalfont\normalsize
   \texttt{\{liuyuang, zuozhi, zhengkai.zhao, lirui.zhang, pan.gao\}@nuaa.edu.cn}} \\[0.35em]
  {\normalfont\normalsize
   \texttt{\{haofeng2, zhengzheliu\}@ln.edu.hk}}
}

\begin{document}

\maketitle

% \begin{figure}[!htbp]
% \centering
% \includegraphics[width=0.95\linewidth]{Section/figures/Pic1.pdf}
% \caption{Qualitative comparison under the $4\times$ upsampling setting. Although regression-based and generative baselines can increase point density, they tend to blur or distort delicate geometric structures such as the thin backrest bars and chair legs. In contrast, our method recovers a more complete and regular structure, producing an upsampled point cloud that is visually closer to the ground truth.}
% \label{fig:pic1}
% \end{figure}
\begin{abstract}
  Point cloud upsampling is essential for 3D applications but remains challenging due to sparse, under-constrained observations. Multi-view images can resolve geometric ambiguity, but existing multi-modal methods do not fully exploit visual cues. We propose \textbf{ControlFlow3D}, a framework that distills knowledge from pretrained foundation models into a lightweight generation network during training, enabling strong performance without multi-view images available at test time. Specifically, our method performs conditional flow matching in a structured latent space defined by a frozen 3D encoder. Multi-view image tokens and 3D latent tokens are jointly clustered into shared prototypes via unsupervised soft KMeans, the resulting prototypes guide the flow trajectory through velocity control and are aligned by a consistency loss that progressively transfers image knowledge into the base network. A Mamba-based geometry decoder then maps latent features to dense coordinates with linear complexity. We also construct ShapeNetPU, a 35K-object multi-modal benchmark that is 31$\times$ larger than prior datasets. Extensive experiments across multiple benchmarks demonstrate the effectiveness of our method and its strong generalization capability. Code and dataset are available at \url{https://github.com/I2-Multimedia-Lab/ControlFlow3D}.
\end{abstract}

\section{Introduction}
Point cloud upsampling aims to generate a denser point set from a sparse input while preserving the underlying 3D surface geometry~\cite{yu2018pu,li2019pu,qian2021pu,he2023grad}. This task is useful for 3D perception and reconstruction systems, where point clouds are widely used in autonomous driving~\cite{lang2019pointpillars,shi2019pointrcnn}, robotics~\cite{qi2017pointnet,qi2017pointnet++,mahler2017dex}, and augmented reality~\cite{newcombe2011kinectfusion,izadi2011kinectfusion}. In practical sensing scenarios, point clouds can be sparse and non-uniform because of limited sensor resolution, occlusions, and viewpoint-dependent sampling. Recovering dense geometry from such observations remains challenging because the sparse input provides only partial measurements of the underlying surface.

\begin{samepage}
The main difficulty of point cloud upsampling is that a sparse input provides incomplete observations of the underlying surface, so multiple dense point sets may be consistent with the same input.
Additional visual observations can provide useful cues for reducing this ambiguity, including object-level semantics, visible boundaries, and geometry-aware latent features~\cite{chen2024mvp,zhao2021sspu,zhang2026utonia}. However, using such cues at inference time can reduce the practicality of an upsampling model, especially when synchronized multi-view images are unavailable. 

\end{samepage}

Existing multi-modal upsampling methods such as MVP-Net~\cite{chen2024mvp} rely on rendered multi-view depth images during both training and inference. Generated by projecting point clouds into 2D views, these depth maps introduce preprocessing and may lose fine 3D structure; they also mainly encode depth and contour rather than richer visual priors from pretrained multi-view image models. SSPU-Net~\cite{zhao2021sspu} avoids test-time images by using differentiable rendering as a self-supervised signal, but image features are never fed to the upsampling network, preventing it from internalizing visual semantics. These limitations motivate our distillation setting: we leverage rich features from pretrained multi-view image models as training-time guidance, while keeping the final model point-only at inference.

\begin{figure}[!htbp]
\centering
\includegraphics[width=1\linewidth]{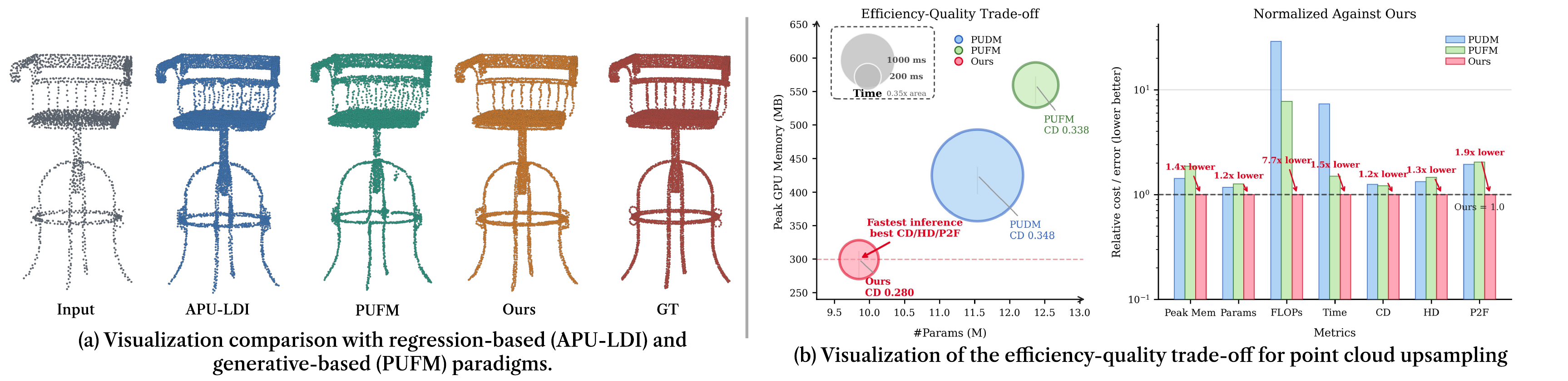}
\caption{Motivation and empirical overview of ControlFlow3D.}
\label{fig:Fig1}
\end{figure}

In this work, we propose \textbf{ControlFlow3D}, a framework that distills knowledge from pretrained multi-view models into an efficient point cloud upsampling network. A central challenge is transferring multi-view knowledge without paired pixel-to-point supervision. We address this through a shared prototype space, where joint clustering of image and 3D tokens discovers cross-modal semantic correspondences in an unsupervised manner.
During training, image patch tokens from VGGT~\cite{wang2025vggt} and 3D target tokens from Utonia~\cite{zhang2026utonia} are projected into a common space and jointly clustered by soft KMeans~\cite{lloyd1982least,dempster1977maximum}. Each cluster yields two complementary prototypes: an image-side prototype that aggregates semantic cues from multi-view observations, and a 3D-side prototype that aggregates dense latent structure at the same query support. A prototype consistency loss aligns the two sides, encouraging the 3D-side prototypes to absorb the texture and semantic cues encoded in the image-side prototypes. The 3D-side prototypes, now enriched with distilled visual knowledge, then produce query-wise scale and shift controls over the latent velocity field via soft query-to-prototype assignment.
Through this design, multi-view images guide the flow trajectory during training, while the base velocity network progressively internalizes the cross-modal knowledge and operates on sparse point clouds alone at inference.

Built on the above core design, ControlFlow3D is composed of three complementary components. First, conditional flow matching is performed in a structured latent space defined by a frozen pretrained 3D encoder, providing a geometry-aware representation that stabilizes generation. 
Second, the prototype-based distillation mechanism injects multi-view knowledge into the velocity field through semantic alignment and control. 
Finally, a Mamba-based geometry decoder maps the generated latent features to dense Euclidean coordinates, enabling efficient high-density reconstruction.

Training this multi-modal pipeline at scale requires precisely aligned 2D and 3D data. 
Existing benchmarks such as PU-GAN~\cite{li2019pu} and PU1K~\cite{qian2021pu} are too small for large-scale multi-modal training, and none provide synchronized multi-view images aligned with point clouds.
We therefore construct \textbf{ShapeNetPU}, a high-fidelity multi-modal dataset derived from ShapeNet, containing 35K objects with clean point clouds and synchronized multi-view images, providing 31$\times$ more data than prior benchmarks.

Extensive experiments demonstrate state-of-the-art (SOTA) performance on multiple benchmarks, with 1.5$\times$ faster inference than flow-matching baselines and 7.3$\times$ faster than diffusion-based methods.
Our method also generalizes well to other upsampling benchmarks~\cite{li2019pu,qian2021pu} and to real-world scene-level point clouds~\cite{dai2017scannet}.
\par
\begin{samepage}
Overall, our main contributions are summarized as follows:
\begin{itemize}
    \item We propose a new \textbf{\textit{paradigm}} for efficient point cloud upsampling that distills knowledge from foundation models into a generation network.

    \item We propose an \textbf{\textit{unsupervised shared-prototype}} distillation mechanism that jointly clusters multi-view image tokens and 3D target tokens into shared centers, making 3D prototypes absorb visual knowledge and produce guidance for the latent velocity field.

    \item We design a \textbf{\textit{Mamba-based}} geometry decoder that efficiently maps generated latent features to dense point coordinates with linear-complexity sequence modeling.

    \item We construct \textbf{\textit{ShapeNetPU}}, a large-scale, high-fidelity multi-modal benchmark for point cloud upsampling, with clean point clouds and synchronized multi-view images.

\end{itemize}
\end{samepage}

\section{Related Work}

\textbf{Flow Matching.}
Flow matching~\cite{lipman2022flow,tong2023improving,liu2022flow} has emerged as a powerful generative paradigm that learns a velocity field transporting a simple prior to the data distribution, offering simulation-free training and efficient ODE-based sampling. It has been successfully applied to high-dimensional visual generation tasks~\cite{esser2024scaling,jin2024pyramidal,polyak2024movie}. We adopt flow matching for 3D upsampling but address a different setting: transporting sparse-conditioned latent features toward dense 3D targets, with multi-view image guidance distilled into a point-only inference model.

\textbf{Point Cloud Upsampling.}
Existing methods fall into regression-based and generative approaches. 
Regression-based methods directly predict dense point coordinates~\cite{yu2018pu,yifan2019patch,qian2021pu,qian2020pugeo,he2023grad, qiu2022pu,rong2024repkpu}.
Although these methods achieve strong benchmark performance, deterministic coordinate regression may favor averaged predictions in ambiguous regions. This reflects the ill-posed nature of point cloud upsampling, where a sparse input may correspond to multiple plausible dense surfaces~\cite{yu2018pu}. Generative methods~\cite{li2019pu,qu2024conditional,zhang2025point,liu2026pufm} model more flexible output distributions, but typically operate in the raw coordinate space, where the lack of structural regularization can cause unstable generation. To reduce the ambiguity of sparse 3D observations, prior work has explored external modalities, primarily for point cloud completion~\cite{dai2017shape,choy20163d,fan2017point,mandikal20183d,wang2018pixel2mesh,zhang2026simba} rather than upsampling.
For upsampling, SSPU-Net~\cite{zhao2021sspu} imposes image-space consistency via differentiable rendering, but uses images only as a pixel-level loss signal without feeding image features into the network. In contrast, MVP-Net~\cite{chen2024mvp} and PULLM~\cite{zhang2025pullm} fuse multi-view depth or language features into the pipeline via concatenation or cross-attention, requiring these external modalities at inference. Our method differs from both: through unsupervised prototype alignment, we distill multi-view knowledge into the generation network during training, allowing the model to operate on point clouds alone at inference. \textbf{More related works are discussed in the Appendix}.

\section{Methodology}

\subsection{Overview}
Given a sparse input point cloud (and multi-view images during training), our goal is to produce a dense and geometrically consistent point cloud.
The framework is organized as follows. We first define a structured latent space using the frozen 3D encoder (\S\ref{sec:latent_repr}). Within this space, we design a \textbf{conditional flow-matching generation network} (\S\ref{sec:flow_matching}) that transports sparse-conditioned features toward the dense prior target. To distill multi-view knowledge into this network, we introduce an \textbf{unsupervised prototype-guided velocity controller} (\S\ref{sec:mv_control}): image and 3D tokens are jointly clustered into shared prototypes, which guide the flow trajectory during training and are progressively absorbed into the base velocity network via a consistency loss. Finally, a \textbf{Mamba-based geometry decoder} (\S\ref{sec:decoder}) maps the generated latent features to dense coordinates,  completing the latent encoding-to-decoding pipeline alongside the frozen encoder.

\begin{figure*}[!t]
\centering
\includegraphics[width=0.98\textwidth]{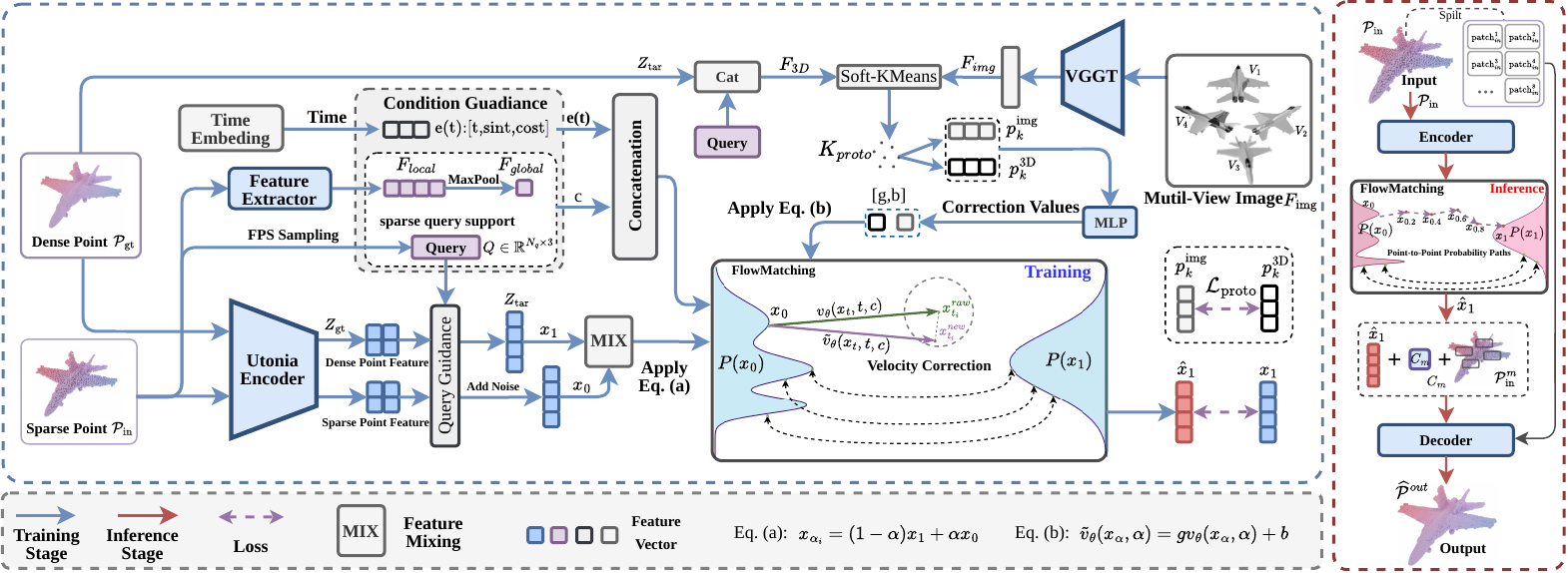}
\caption{Overview of our Latent Flow Matching. We first use a frozen pretrained 3D encoder (Utonia) to define a structured target latent space for dense geometry. Within this space, we design a \textbf{conditional flow-matching generation network} that transports sparse-conditioned features toward the dense prior target. To distill multi-view knowledge into this network, we introduce an \textbf{unsupervised prototype-guided velocity controller}  that jointly clusters image $F_{img}$ and 3D tokens $F_{3D}$ into shared prototypes. During inference, the model takes only the sparse point cloud as input and produces the final dense point cloud.}
\label{fig:method_Pipeline}
\end{figure*}

\subsection{Structured 3D Latent Representation}
\label{sec:latent_repr}

Point clouds are inherently unstructured and irregular, making direct generation in Euclidean coordinate space unstable and difficult to regularize. 
To address this, we shift the generative process to a structured latent manifold by adopting a frozen pretrained 3D encoder to define the target latent space. We denote the sparse input point cloud as $\mathcal{P}_{\mathrm{in}}$ and the dense ground-truth point cloud as $\mathcal{P}_{\mathrm{gt}}$. The sparse query support is sampled from the input as $Q\in \mathbb{R}^{N_q \times 3} = \mathrm{FPS}(\mathcal{P}_{\mathrm{in}}),
$
where $N_q$ is the number of sparse query points, set to 256 by default.
The encoder extracts latent representations from the dense ground-truth point clouds:
$
    Z_{\mathrm{gt}} = E_{\mathrm{3D}}(\mathcal{P}_{\mathrm{gt}}),
    Z_{\mathrm{gt}} \in \mathbb{R}^{N_g \times D},
$
where $E_{\mathrm{3D}}$ denotes Utonia~\cite{zhang2026utonia}, $N_g$ is the number of latent tokens, and $D$ is the latent dimension.
These frozen features provide a geometry-aware but coordinate-abstract target space, allowing the generation network to learn dense 3D structure in latent space before coordinate decoding.
To align this dense 3D prior with the sparse query support used by the latent flow, we interpolate the dense latent features onto $Q$:
    $
    Z_{\mathrm{tar}}
    =\mathrm{Interp}(Q;\,\mathcal{P}_{\mathrm{gt}},\,Z_{\mathrm{gt}}),
    Z_{\mathrm{tar}} \in \mathbb{R}^{N_q \times D}.
    $
Here, $\mathrm{Interp}$ denotes 3-NN feature interpolation, which estimates a dense latent feature at each query location in $Q$. Here, $Q$ defines the spatial support for latent flow matching, and $Z_{\mathrm{tar}}$ serves as the target endpoint in the flow process. The encoder is kept frozen throughout training to preserve the geometric priors from large-scale pretraining and provide a stable target distribution for flow matching. This subsection defines the \emph{target space} only; the generation network that operates within it and the multi-view distillation mechanism are described in the following subsection.

\subsection{Multi-View Distilled Latent Flow Matching}
\label{sec:flow_matching}
This section presents the core generation network: a conditional latent flow that transports sparse-aligned features toward the 3D target, with multi-view knowledge distilled into the velocity network during training to enable image-free inference.

\textbf{Sparse geometry conditioning.}
The sparse input point cloud serves as the geometric condition for latent transport. We encode the sparse observation $\mathcal{P}_{\mathrm{in}}$ using a lightweight PointNet++~\cite{qi2017pointnet++} backbone $\mathrm{FE}(\cdot)$ to obtain point-wise local features and a global shape descriptor:
    $
    F_{\mathrm{local}} = \mathrm{FE}(\mathcal{P}_{\mathrm{in}},Q),
    F_{\mathrm{global}} = \operatorname{MaxPool}(F_{\mathrm{local}}).
    $
The sparse-query support $Q$ anchors all subsequent feature operations to explicit 3D locations.
The local sparse features are aligned to $Q$, and the conditioning vector is then formed as $c = \operatorname{Concat}[F_{\mathrm{local}},\, F_{\mathrm{global}},\, Q]$, ensuring that the latent dynamics remain continuously grounded in the observed sparse geometry. 

\textbf{Conditional Latent Flow Matching.}
With the sparse-query support $Q$ and condition $c$ already defined above, we construct the source state $x_0 = \mathrm{Gather}(E_{\mathrm{3D}}(\mathcal{P}_{\mathrm{in}}),Q) + \sigma\epsilon,$ where $\epsilon \sim \mathcal{N}(0,I)$ denotes Gaussian noise, and $\sigma$ controls the noise strength. Here, $\mathrm{Gather}(\cdot)$ reuses the FPS indices used to sample $Q$ to select the corresponding sparse latent tokens from $E_{\mathrm{3D}}(\mathcal{P}_{\mathrm{in}})$, yielding $N_q$ tokens of dimension $D$. In the implementation, Gaussian noise is added to the gathered features, followed by 3-NN interpolation onto the sparse-query support $Q$. The target endpoint is $x_1 = Z_{\mathrm{tar}}$. Although both $x_0$ and $x_1$ are encoded by $E_{\mathrm{3D}}$, they carry fundamentally different information: $x_0$ captures only sparse-resolution context at the query locations, while $x_1$ encodes dense-resolution geometric detail indexed to the same spatial support. The flow must therefore learn to synthesize the missing fine-grained structural information that the sparse encoding lacks. Given these two endpoints, we define a linear interpolation path in latent space: $x_t = t\,x_1 + (1 - t)\,x_0,$ where $t$ is sampled with cosine reparameterization to distribute supervision over a smoother transport trajectory. A velocity network $\widetilde{v}_\theta$ is trained to predict the displacement from $x_0$ to $x_1$ at each interpolant $x_t$:
\begin{equation}
\label{eq:fm_loss}
    \mathcal{L}_{\mathrm{FM}}
    =
    \mathbb{E}_{t,\,x_0,\,x_1}
    \!\left[
    \left\|
    \widetilde{v}_\theta(x_t,\,t,\,c,\,\mathcal{I})
    -
    (x_1 - x_0)
    \right\|_2^2
    \right],
\end{equation}
where $c$ denotes the sparse geometric condition and $\mathcal{I}$ denotes the multi-view image features extracted by VGGT~\cite{wang2025vggt}. This formulation casts point cloud upsampling as a conditional transport problem on a structured latent manifold: the flow learns geometry-aware structural evolution, while the recovery of Euclidean coordinates is deferred to the subsequent decoding stage. \textbf{More details are provided in the Appendix}.

\textbf{Unsupervised Prototype-Guided Velocity Control}
\label{sec:mv_control}
    
While $Z_{\mathrm{tar}}$ defines the target latent endpoint, the sparse condition $c$ alone may leave the transport trajectory ambiguous. Multi-view images carry rich complementary cues, such as appearance, boundaries and part-level semantics, that can disambiguate this trajectory. Rather than directly fusing 2D features into the 3D pipeline, we design a distillation mechanism based on unsupervised prototype learning: image knowledge guides the velocity field during training and is progressively absorbed into the base network, enabling image-free inference. To transfer multi-view cues without pixel-to-point correspondence while retaining point-only inference, we use shared prototypes as a compact interface for training-time velocity guidance.
 
Given $\mathcal{I}$, VGGT~\cite{wang2025vggt} extracts multi-view patch tokens $F_{\mathrm{img}}\in\mathbb{R}^{V\times T\times C_I}$, where $V$, $T$, $C_I$ denote the number of views, tokens per view, and feature dimension, respectively. Each view is optionally average-pooled to $T'$ tokens and flattened into $F_{\mathrm{mv}}\in\mathbb{R}^{(VT')\times C_I}$. Meanwhile, target-side 3D tokens are formed as $F_{\mathrm{3D}}=\mathrm{Concat}[Z_{\mathrm{tar}},\,Q]$ from the target latent features and query coordinates defined in \S\ref{sec:latent_repr}. Both token sets are projected to a shared space, where soft-KMeans~\cite{lloyd1982least,dempster1977maximum} produces $K_{proto}$ shared centers without any label supervision. All objects share the same learned prototype seeds. For each object, soft-KMeans updates the centers from these seeds while preserving their indices, providing a common index convention across objects. The image-side prototypes $\mathbf{p}^{\mathrm{img}}_k$ and 3D-side prototypes $\mathbf{p}^{\mathrm{3D}}_k$ are obtained by softly aggregating their respective tokens around the same updated centers.
 
To inject prototype-level guidance into the latent flow, we compute a prototype distribution for each query. Let $\mathbf{s}_k$ denote the $k$-th shared center, and let $q_j\in Q$ be the coordinate of the $j$-th query. We form the query token $u_j=\mathrm{Concat}[x_{t,j},\,q_j,\,e(t)]$, where $e(t)=[t,\sin t,\cos t]$, and compute
\begin{equation}
    a_{k j}
    =
    \frac{
    \exp\left(\tau\,\langle \psi_q(u_j),\,\psi_s(\mathbf{s}_k)\rangle\right)}
    {\sum_{\ell=1}^{K_{proto}}
    \exp\left(\tau\,\langle \psi_q(u_j),\,\psi_s(\mathbf{s}_\ell)\rangle\right)},
\end{equation}
where $\psi_q$ and $\psi_s$ are learnable projections, $\tau$ controls the assignment sharpness, and $\sum_k a_{kj}=1$. A lightweight MLP predicts bounded prototype-level controls $\mathbf{g}^{proto}_k$ and $\mathbf{b}^{proto}_k$ from the image-side prototypes and pooled multi-view context. Query-wise controls are then obtained by assignment-weighted aggregation:
\begin{equation}
    \mathbf{g}_{:,j}
    =
    \sum_{k=1}^{K_{proto}} a_{kj} \mathbf{g}^{proto}_k,
    \qquad
    \mathbf{b}_{:,j}
    =
    \sum_{k=1}^{K_{proto}} a_{kj} \mathbf{b}^{proto}_k .
\end{equation}
The controlled velocity is: 
\begin{equation}
\label{eq:affine}
    \widetilde{v}_\theta(x_t,\,t,\,c,\,\mathcal{I})
    =
    \mathbf{g}(x_t,\,t,\,\mathcal{I})
    \odot
    v_\theta(x_t,\,t,\,c)
    +
    \mathbf{b}(x_t,\,t,\,\mathcal{I}),
\end{equation}
where $v_\theta(x_t,t,c)$ is the base velocity from the sparse condition, $\mathbf{g}\in\mathbb{R}^{D\times N_q}$ is a query-wise scaling term centered around one that reweights the transport direction, and $\mathbf{b}\in\mathbb{R}^{D\times N_q}$ is a bounded additive correction from the multi-view prototypes. The soft assignment transfers prototype-level semantic guidance to spatially anchored query locations.
 
To keep the visual controller aligned with the 3D target structure, we impose a prototype consistency regularization:
\begin{equation}
\label{eq:proto_loss}
    \mathcal{L}_{\mathrm{proto}}
    =
    1
    -
    \frac{1}{K_{proto}}
    \sum_{i=1}^{K_{proto}}
    \cos\!\left(\mathbf{p}^{\mathrm{img}}_i,\;\mathbf{p}^{\mathrm{3D}}_i\right).
\end{equation}

\providecommand{\boldsymbol}[1]{\mathbf{#1}}

\subsection{Geometry Decoder}
\label{sec:decoder}
\begin{figure}[!htbp]
\centering
\includegraphics[width=1\linewidth]{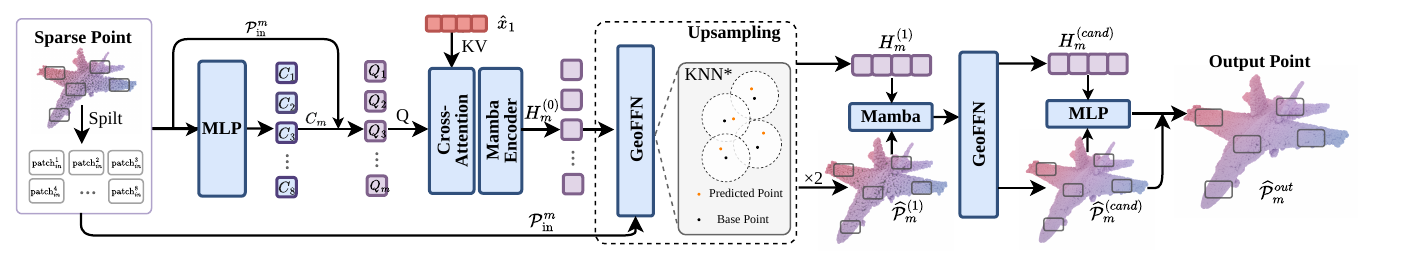}
\caption{Architecture of the Patch-wise Mamba-based Geometry Decoder. The decoder fuses local patch features with the terminal latent endpoint $\widehat{x}_1$, then applies cascaded GeoFFN and Mamba blocks to predict dense patches with supervision on intermediate and final outputs.
}
\label{fig:Pipeline2}
\end{figure}

The geometry decoder maps the latent features to Euclidean coordinates. The density growth inherent in upsampling makes this stage computationally demanding. We therefore adopt a SSM-based architecture~\cite{gu2023mamba}, whose linear-complexity selective sequence modeling balances efficiency with long-range modeling capacity. Figure~\ref{fig:Pipeline2} illustrates the overview of Mamba-based decoder.

\textbf{Latent Endpoint Estimation.}
Let $\widehat{x}_1$ denote the terminal latent feature passed to the decoder. During training, we first estimate the latent endpoint from the sampled state on the linear interpolation path and the predicted velocity:
\begin{equation}
    \widehat{x}_1
    =
    x_t + (1 - t)\,\widetilde{v}_\theta(x_t,\,t,\,c,\,\mathcal{I}).
\end{equation}

\textbf{Patch-wise Feature Fusion.}
The sparse input $\mathcal{P}_{\mathrm{in}}$ is divided into patches $\{\mathcal{P}_{\mathrm{in}}^m\}_{m=1}^{M}$. Let $Q$ be the sparse-query support used by the latent flow, and let $C^m$ denote the point-wise sparse local condition gathered for patch $m$. For each patch, the predicted latent tokens $\widehat{x}_1$ are interpolated to the patch points via 3-nearest-neighbor weighting on the query support $Q$, yielding a patch-specific flow condition that is then fused with $C^m$ through the first decoder block. We use CrossAttention to fuse the features to obtain the input features for the Patch-wise Up-sampling module: $H^{(0)}_m = CA(\mathcal{P}_{\mathrm{in}}^m,\;\widehat{x}_1,\;C^m)$.

\textbf{Geometry-Aware Mamba Decoder.}
We serialize patch points with space-filling curve orders and aggregate bidirectional scans to reduce the anisotropic bias of one-dimensional ordering. Block~1 uses Hilbert and trans-Hilbert orders to preserve local neighborhoods during the initial fusion of sparse conditions and predicted latent features. Block~2 uses $z$ order and trans-$z$ order at higher point density, providing complementary traversal for refinement. Since linear scans may weaken geometric cues, we adopt GeoFFN, inspired by the LCFFN module in LCM~\cite{zha2024lcm}, to enhance geometry awareness. The two-block decoder produces patch point sets and hidden features:
\begin{equation}
    \begin{aligned}
    (\widehat{\mathcal{P}}^{(1)}_m,\;H^{(1)}_m)
    &=
    GeoFFN_1(MambaBlock_1(\mathcal{P}_{\mathrm{in}}^m,\;H^{(0)}_m)),
    \\
    (\widehat{\mathcal{P}}^{(2)}_m,\;H^{(2)}_m)
    &=
    GeoFFN_2(MambaBlock_2(\widehat{\mathcal{P}}^{(1)}_m,\;H^{(1)}_m)).
    \end{aligned}
\end{equation}
The Block-1 points and the Block-2 branch are concatenated into candidate points $\widehat{\mathcal{P}}^{\mathrm{cand}}_m = [\widehat{\mathcal{P}}^{(1)}_m;\widehat{\mathcal{P}}^{(2)}_m]$ and features $H^{\mathrm{cand}}_m = [H^{(1)}_m;H^{(2)}_m]$, which are processed by a position-aware residual head that predicts a bounded coordinate offset:
\begin{equation}
    \widehat{\mathcal{P}}^{out}_m
    =
    \widehat{\mathcal{P}}^{\mathrm{cand}}_m
    +
    \tanh
    \left(
    MLP\!\left(
    [H^{\mathrm{cand}}_m,\;\mathrm{PE}(\widehat{\mathcal{P}}^{\mathrm{cand}}_m)]
    \right)
    \right).
\end{equation}
Here, $[\cdot;\cdot]$ denotes concatenation along the point/token dimension.
The final dense point cloud $\widehat{\mathcal{P}}$ is obtained by merging all patch outputs. Importantly, the decoder is a deterministic Euclidean reconstruction module conditioned on the predicted latent features and does not introduce another generative process. $PE$ denotes positional encoding.
The geometric loss is applied in a stage-wise manner to the decoder outputs:
\begin{equation}
    \mathcal{L}_{\mathrm{geo}}
    =
    \sum_{l}
    CD(\widehat{\mathcal{P}}^{(l)},\;\mathcal{P}_{\mathrm{gt}})
    +
    \sum_{l}
    \frac{1}{M} \sum_{m}^MCD(\widehat{\mathcal{P}}_m^{(l)},\;\mathcal{P}^{gt}_{\mathrm{m}})
\end{equation}
where $\widehat{\mathcal{P}}^{(l)} = [\widehat{\mathcal{P}}_m^{(l)}]_{m=1}^M,$ denotes the output of decoder block $l$, $\mathcal{P}^{gt}_{\mathrm{m}}$ is also divided into $M$ parts, ${CD}(\cdot,\cdot)$ is the Chamfer Distance.

\subsection{Overall Training Objective}
\label{sec:training}

The full framework is optimized by combining latent flow matching, prototype alignment, and coordinate reconstruction.  The overall objective is:

\begin{equation}
    \mathcal{L}
    =
    \mathcal{L}_{\mathrm{geo}}
    +
    \lambda_{\mathrm{FM}}\,\mathcal{L}_{\mathrm{FM}}
    +
    \lambda_{\mathrm{proto}}\,\mathcal{L}_{\mathrm{proto}}.
\end{equation}
Each loss term reflects the separation of concerns in our framework: $\mathcal{L}_{\mathrm{FM}}$ drives the latent trajectory toward the Utonia 3D prior target, $\mathcal{L}_{\mathrm{proto}}$ aligns the multi-view prototypes with the 3D prior to keep the visual controller geometrically consistent, and $\mathcal{L}_{\mathrm{geo}}$ supervises the final coordinate reconstruction.

\section{Experiments}
\label{exp}
\subsection{Datasets and Metrics}

Our main evaluation is conducted on \textbf{ShapeNetPU}, a multi-modal point cloud upsampling benchmark. It contains 35K valid samples from 13 categories. Each object is associated with eight selected views. We also evaluate on PU-GAN~\cite{li2019pu} and PU1K~\cite{qian2021pu}. Following prior works~\cite{qian2021pu,he2023grad,li2024learning}, we report Chamfer Distance (CD), Hausdorff Distance (HD), and point-to-surface distance (P2F), all in units of $10^{-3}$. Lower is better for all three metrics. For more details of the datasets, please refer to the Appendix.

\begin{table}[!b]
\centering
\begin{minipage}[t]{0.49\textwidth}
\vspace{0pt}
\centering
\setlength{\tabcolsep}{2pt}
\captionof{table}{Quantitative results on ShapeNetPU at $4\times$ and $16\times$ upsampling. The top three results are in \first{red}, \second{blue}, and \third{green}, respectively.}
\label{tab:quantitative_comparison}
\footnotesize
\begin{tabular*}{\linewidth}{@{\extracolsep{\fill}}lcccccc@{}}
\toprule
\multirow{2}{*}{Method}
& \multicolumn{3}{c}{ShapeNetPU $4\times$}
& \multicolumn{3}{c}{ShapeNetPU $16\times$} \\
\cmidrule(lr){2-4} \cmidrule(lr){5-7}
& CD$\downarrow$ & HD$\downarrow$ & P2F$\downarrow$
& CD$\downarrow$ & HD$\downarrow$ & P2F$\downarrow$ \\
\midrule
PU-Net~\cite{yu2018pu}        & 1.084 & 12.628 & 5.381 & 1.274 & 20.292 & 6.894 \\
PU-GAN~\cite{li2019pu}        & 0.645 & 8.829  & 3.420 & 0.459 & 13.494 & 3.852 \\
PU-GCN~\cite{qian2021pu}      & 0.593 & 7.692  & 2.527 & 0.312 & 9.603  & 2.722 \\
Grad-PU~\cite{he2023grad}     & 0.414 & 3.429  & 1.684 & 0.280 & 3.372  & 1.892 \\
APU-LDI~\cite{li2024learning} & 0.364 & \third{3.182}  & \second{1.447} & \second{0.246} & \second{2.735}  & \third{1.845} \\
PUDM~\cite{qu2024conditional} & \third{0.348} & \second{3.180} & \third{1.590} & 0.278 & \third{2.952}  & \second{1.757} \\
PUFM~\cite{liu2026pufm}       & \second{0.338} & 3.286 & 1.672  & \third{0.265} & 3.215  & 1.903 \\
\midrule
Ours                          & \first{0.280} & \first{2.259} & \first{0.818} & \first{0.221} & \first{2.498} & \first{1.167} \\
\bottomrule
\end{tabular*}
\end{minipage}
\hfill
\begin{minipage}[t]{0.49\textwidth}
\vspace{0pt}
\centering
\setlength{\tabcolsep}{2pt}
\captionof{table}{Quantitative results on PU-GAN and PU1K at $4\times$ upsampling. The top three results are in \first{red}, \second{blue}, and \third{green}, respectively.}
\label{tab:pugan_pu1k}
\footnotesize
\begin{tabular*}{\linewidth}{@{\extracolsep{\fill}}lcccccc@{}}
\toprule
\multirow{2}{*}{Method}
& \multicolumn{3}{c}{PU-GAN $4\times$}
& \multicolumn{3}{c}{PU1K $4\times$} \\
\cmidrule(lr){2-4} \cmidrule(lr){5-7}
& CD$\downarrow$ & HD$\downarrow$ & P2F$\downarrow$
& CD$\downarrow$ & HD$\downarrow$ & P2F$\downarrow$ \\
\midrule
PU-Net~\cite{yu2018pu}        & 0.507 & 4.312 & 4.694 & 1.155 & 15.170 & 4.834 \\
PU-GAN~\cite{li2019pu}        & 0.284 & 3.301 & 2.660 & 0.735 & 9.939  & 4.234 \\
PU-GCN~\cite{qian2021pu}      & 0.274 & 2.974 & 2.535 & 0.585 & 7.577  & 2.499 \\
Grad-PU~\cite{he2023grad}     & 0.260 & 2.462 & \third{1.949} & \third{0.404} & \third{3.732} & \third{1.474} \\
APU-LDI~\cite{li2024learning} & \first{0.232} & \first{1.673} & \first{1.507} & \second{0.399} & \second{3.603} & \second{1.287} \\
PUDM~\cite{qu2024conditional} & 0.264 & 2.441 & 3.080 & 0.478 & 4.441  & 1.725 \\
PUFM~\cite{liu2026pufm}       & \second{0.250} & \second{2.341} & 2.292 & 0.421 & 3.763  & 1.682 \\
\midrule
Ours                          & \third{0.257} & \third{2.418} & \second{1.854}
                              & \first{0.390} & \first{3.492} & \first{1.222} \\
\bottomrule
\end{tabular*}
\end{minipage}
\end{table}

\subsection{Quantitative and Qualitative Comparison on ShapeNetPU}

\textbf{Quantitative Comparison.}

Table~\ref{tab:quantitative_comparison} compares our method with representative regression-based and generative baselines on ShapeNetPU under $4\times$ and $16\times$
 upsampling. For a fair comparison, we have replicated all the methods based on our experimental setup and on the dataset. Our method achieves the best results on all three metrics under both settings. Two comparisons are particularly informative. First, our method outperforms PUFM, the strongest coordinate-space flow matching baseline, supporting our claim that latent-space generation is more stable than coordinate-space generation. Second, the largest gain appears in P2F, which measures distance to the underlying continuous surface rather than to discrete target points. The roughly 50\% improvement over competing methods suggests that the 3D latent space provides a strong surface-aware representation, while multi-view guidance further helps the generated points align with the underlying geometric manifold.

\begin{figure}[!htbp]
\centering
\includegraphics[width=0.95\linewidth]{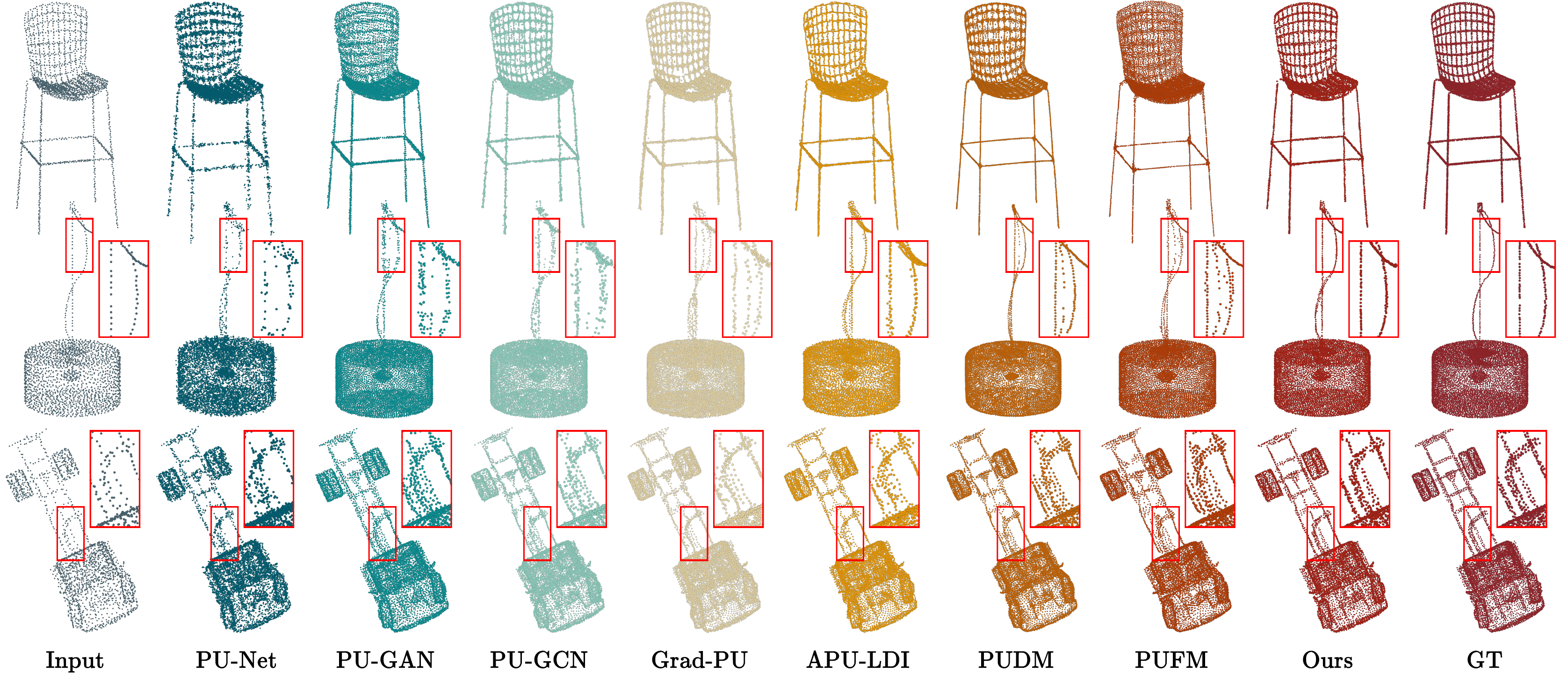}
\caption{Qualitative visualization on ShapeNetPU under the $4\times$ upsampling setting.}
\label{fig:shapenetpu_vis}
\end{figure}

\textbf{Qualitative Comparison.}
Figure~\ref{fig:shapenetpu_vis} presents qualitative $4\times$ upsampling results on ShapeNetPU. Compared with the sparse input, the generated point clouds recover denser and more continuous object surfaces while preserving local structures such as thin parts, boundaries, and curved regions. These visual results are consistent with the quantitative gains in Table~\ref{tab:quantitative_comparison}, especially the lower P2F error, showing that the predicted points are better attached to the underlying surface instead of simply increasing point density. It demonstrates the effectiveness of our proposed distillation mechanism.

\subsection{Quantitative and Qualitative Comparison on PU-GAN and PU1K}

\begin{figure}[!hbp]
\centering
\includegraphics[width=1\linewidth]{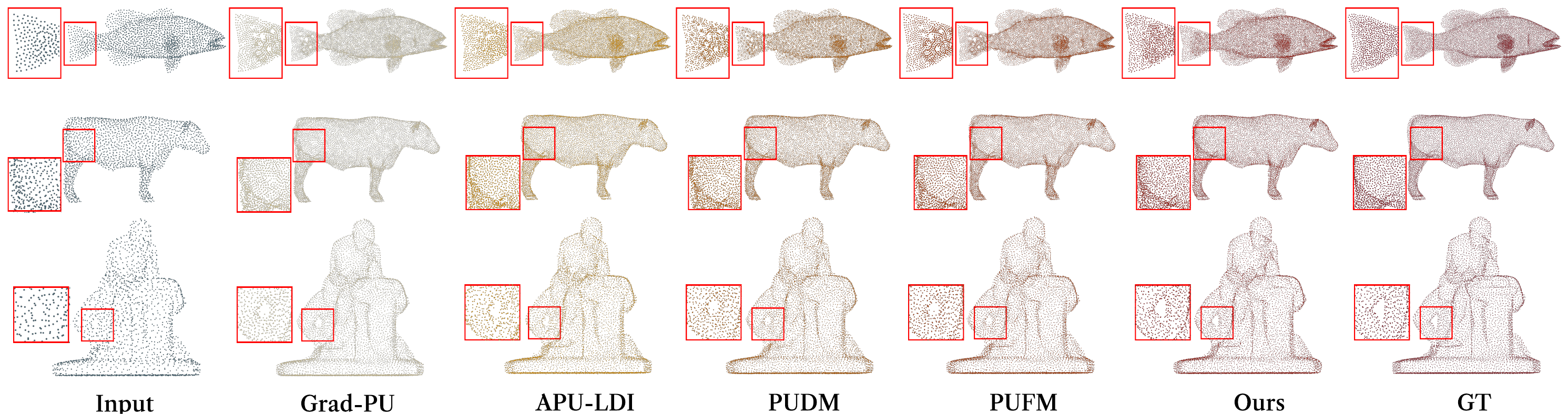}
\caption{Qualitative visualization on PU-GAN under the $4\times$ upsampling setting.}
\label{fig:pugan_vis}
\end{figure}

To evaluate the transferability of our method, we further conduct experiments on the PU-GAN and PU1K benchmarks to assess cross-dataset generalization. \textbf{Note}: PU-GAN and PU1K use overlapping local patches as input samples, whereas our ShapeNetPU pre-training uses complete object-level point clouds; the non-overlapping patches in our framework are only used inside the decoder. Thus, PU-GAN/PU1K provide less geometric and semantic context to the frozen 3D encoder, and their lack of aligned multi-view images further weakens the benefit of visual distillation.
Cropped patches provide incomplete geometry, making it harder for the frozen 3D encoder to extract sufficiently informative latent features, while the absence of images also prevents the prototype controller from fully exploiting visual guidance.
These factors explain the limited gains on PU-GAN as shown in Tab~\ref{tab:pugan_pu1k}.
PU1K shows stronger results partly because it contains PU-GAN and ShapeNet-derived shapes, making its distribution closer to ShapeNetPU; after removing all PU1K test shapes from the ShapeNetPU pre-training split, this reflects distributional overlap rather than test-shape leakage. Although our PU-GAN metrics are not uniformly state of the art, Figure~\ref{fig:pugan_vis} shows that our visual details, such as local structures and surface continuity, are comparable to recent SOTA methods. This indicates that the learned information from ShapeNetPU remains transferable despite patch-based inputs and missing multi-view supervision.

\begin{figure}[!htbp]
\centering
\includegraphics[width=0.95\linewidth]{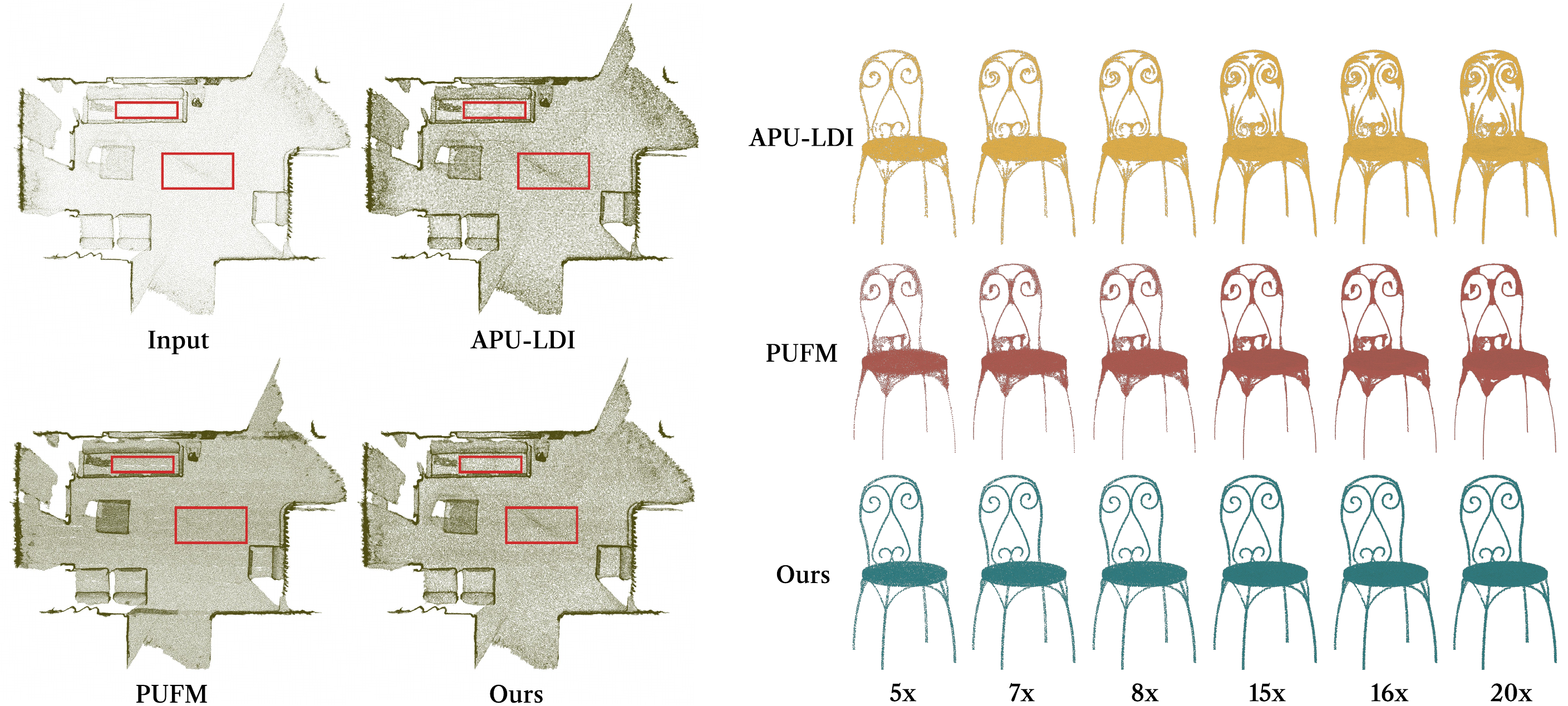}
\caption{More visualization results. (a) Visualization of ScanNet; (b) Visualization of Arbitrary-Rate Upsampling.}
\label{fig:ScanNetArbitrary}
\end{figure}

\subsection{Arbitrary-Rate Upsampling}
A practical upsampler should not be tied to a single density ratio. We therefore evaluate arbitrary-rate upsampling on ShapeNetPU and compare with PUFM and APU-LDI. This experiment tests whether the learned transport remains stable as the target density changes. Before reporting the numerical comparison, Figure~\ref{fig:ScanNetArbitrary}.(b) provides a visual example of how different methods behave when the requested output density varies.
The stable results across ratios suggest that image guidance during training remains reliable at different output resolutions, with the distilled network retaining 2D detail cues. \textbf{We have provided quantitative results in the Appendix.}

\subsection{Ablation Study}
We conduct ablation studies on ShapeNetPU under the $4\times$ setting to demonstrate the effectiveness of our method.

\begin{table*}[!tbp]
\centering
\caption{Ablation studies on ShapeNetPU under $4\times$ upsampling. Lower is better.}
\label{tab:ablation_main}
\footnotesize
\setlength{\tabcolsep}{3pt}
\begin{minipage}[t]{0.49\textwidth}
\centering
\begin{tabular*}{\linewidth}{@{\extracolsep{\fill}}lccc}
\toprule
Variant & CD$\downarrow$ & HD$\downarrow$ & P2F$\downarrow$ \\
\midrule
\multicolumn{4}{c}{\textit{Image Distillation}} \\
w/o distillation & 0.329 & 2.842 & 1.139 \\
w/ distillation & \textbf{0.280} & \textbf{2.259} & \textbf{0.818} \\
\midrule
\multicolumn{4}{c}{\textit{2D Backbone}} \\
DINOv2~\cite{oquab2023dinov2} & 0.292 & 2.374 & 0.884 \\
VGGT~\cite{wang2025vggt} & \textbf{0.280} & \textbf{2.259} & \textbf{0.818} \\
\midrule
\multicolumn{4}{c}{\textit{Prototype Count}} \\
$K_{proto}=2$ & 0.302 & 2.493 & 0.952 \\
$K_{proto}=4$ & 0.291 & 2.378 & 0.882 \\
$K_{proto}=6$ & 0.284 & 2.301 & 0.841 \\
$K_{proto}=10$ & 0.283 & 2.287 & 0.834 \\
$K_{proto}=8$ (default) & \textbf{0.280} & \textbf{2.259} & \textbf{0.818} \\
\bottomrule
\end{tabular*}
\end{minipage}%
\hfill
\begin{minipage}[t]{0.49\textwidth}
\centering
\begin{tabular*}{\linewidth}{@{\extracolsep{\fill}}lccc}
\toprule
Variant & CD$\downarrow$ & HD$\downarrow$ & P2F$\downarrow$ \\
\midrule
\multicolumn{4}{c}{\textit{GeoFFN}} \\
w/o GeoFFN & 0.299 & 2.454 & 0.920 \\
w/ GeoFFN & \textbf{0.280} & \textbf{2.259} & \textbf{0.818} \\
\midrule
\multicolumn{4}{c}{\textit{3D Encoder}} \\
PTv3~\cite{wu2024point} & 0.287 & 2.361 & 0.878 \\
Utonia~\cite{zhang2026utonia} & \textbf{0.280} & \textbf{2.259} & \textbf{0.818} \\
\midrule
\multicolumn{4}{c}{\textit{Mamba Scan}} \\
Sequential scan & 0.306 & 2.514 & 0.962 \\
Hilbert only & 0.294 & 2.401 & 0.890 \\
$z$-order only & 0.298 & 2.436 & 0.911 \\
Bi-dir. same scan & 0.288 & 2.352 & 0.861 \\
Hilbert/rev. + $z$/rev. & \textbf{0.280} & \textbf{2.259} & \textbf{0.818} \\
\bottomrule
\end{tabular*}
\end{minipage}
\end{table*}

\begin{figure}[!htbp]
\centering
\includegraphics[width=0.98\linewidth]{\detokenize{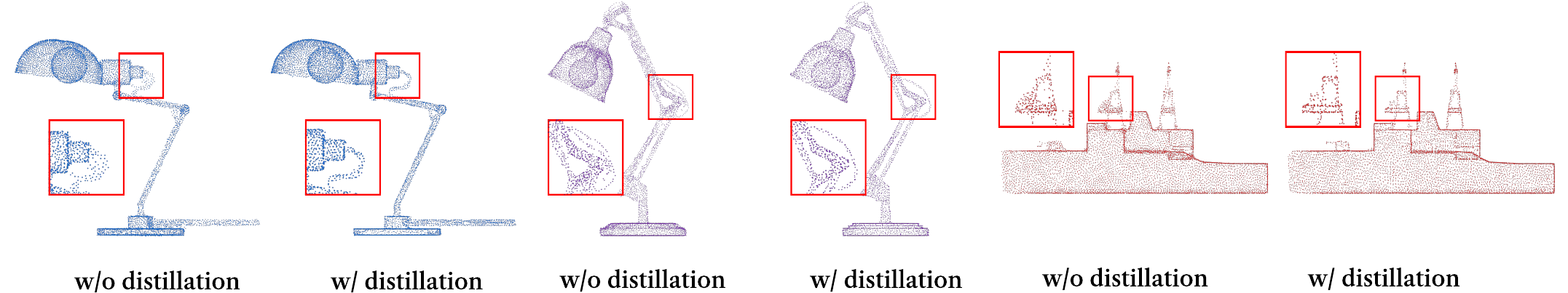}}
\caption{Qualitative comparison with and without image distillation on ShapeNetPU under the $4\times$ upsampling setting.}
\label{fig:mvd_ablation}
\end{figure}

\begin{figure}[!htbp]
\centering
\includegraphics[width=\linewidth]{\detokenize{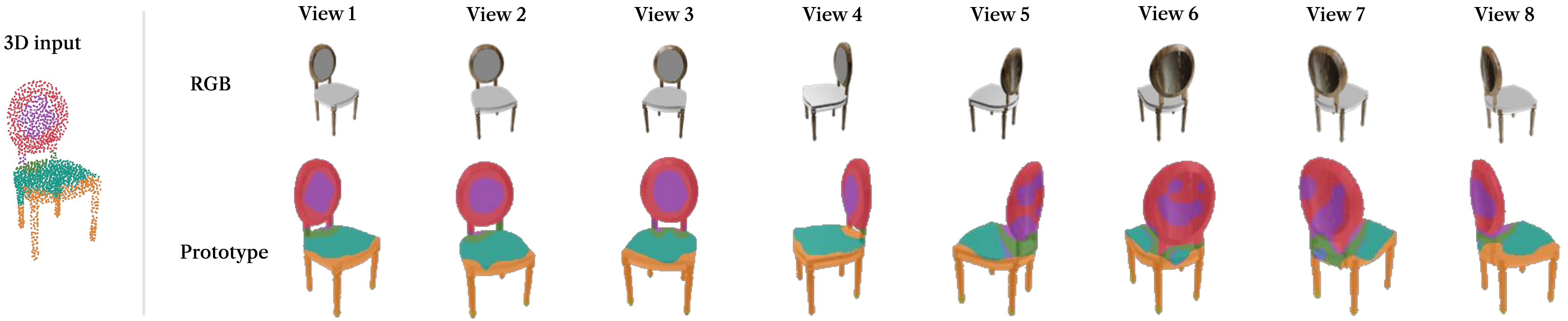}}
\caption{Shared-prototype visualization of a chair: colored 3D input (left), eight RGB views (top right), and assignment maps (bottom right). Colors denote shared prototype indices.}
\label{fig:shared_proto_vis}
\end{figure}

Table~\ref{tab:ablation_main} shows that the gains do not come from a single isolated component. Removing image distillation weakens the training signal that connects sparse geometry to dense surface structure, as also shown by the visual comparison in Figure~\ref{fig:mvd_ablation}. Image distillation improves CD, HD, and P2F while retaining point-only inference. Figure~\ref{fig:shared_proto_vis} visualizes corresponding prototype assignments across the 3D input and image views.

Using VGGT as the 2D backbone outperforms DINOv2, indicating that stronger multi-view visual tokens better guide latent transport. In the 3D space, a larger-scale pretrained encoder further improves performance. This is expected because the frozen 3D encoder defines the target latent space, and its representation capacity affects latent supervision quality. Nevertheless, the overall trend remains stable as long as the encoder provides an informative geometry-aware latent space.

For the prototype count, increasing $K_{proto}$ from 2 to 8 consistently improves all metrics, while 10 prototypes bring no further gain, suggesting that too few prototypes under-represent multi-view semantics, whereas excessive clustering may fragment the shared prototype space.

GeoFFN improves over the version without it, showing that lightweight geometry-aware decoding helps predicted points stay attached to the surface. The lower CD and HD complement the improvement in P2F, indicating more accurate coordinate reconstruction from the latent features. This is consistent with the role of GeoFFN in enhancing geometry awareness within the Mamba-based geometry decoder when predicting dense point coordinates. For Mamba scan design, spatial scan curves improve geometric fidelity over naive sequential ordering, and complementary forward/reverse scans achieve the best result. Together, these results support the overall design: the latent generator provides structured features, while the Mamba with GeoFFN decoder converts them into dense and faithful coordinates. \textbf{More analyses are provided in Appendix.}

\subsection{Generalization to Real-World Point Cloud Upsampling in ScanNet}
Finally, we evaluate qualitative generalization on ScanNet~\cite{dai2017scannet}, a real-world indoor RGB-D scanning dataset. 
Figure~\ref{fig:ScanNetArbitrary}(a) shows representative ScanNet upsampling results. Our method increases point density along visible walls, floors, and object boundaries while largely preserving the layout of the original scan. 
This suggests that the frozen Utonia latent space provides stable representations at the scene level, while the Mamba decoder with GeoFFN helps a compact network transfer pretrained 3D knowledge to real scans.

\section{Limitation and Conclusion}
\label{limitationandconclusion}
\textbf{Limitation.}
Our framework depends on a frozen pretrained 3D encoder and thus inherits the expressiveness of its latent space.  When the input is an incomplete local patch rather than a complete object-level point cloud, the encoder may lack sufficient geometric and semantic context, and out-of-distribution shapes may not be faithfully represented by the pretrained 3D encoder. Although inference is image-free, training still requires aligned multi-view images, limiting scalability to datasets with such pairs.

\textbf{Conclusion.}
We presented \textbf{ControlFlow3D}, a multi-view distilled latent flow matching framework for point cloud upsampling. Our method shifts generation from unstable raw coordinates to a structured latent space defined by a pretrained 3D encoder, and uses an unsupervised shared-prototype distillation mechanism to transfer multi-view visual knowledge into the velocity network during training. A Mamba-based geometry decoder maps the generated latent features to dense coordinates with linear-complexity sequence modeling. We further construct ShapeNetPU, a multi-modal upsampling dataset for training. Extensive experiments on multiple benchmarks demonstrate the effectiveness and strong generalization of our method.

\clearpage
\begin{ack}
This work was supported by the National Natural Science Foundation of China (No.~62272227).
The authors declare no competing interests.
\end{ack}

%\section*{References}
\bibliographystyle{unsrt}  %plainnat,abbrvnat,unsrtnat
\small
\bibliography{Reference}
\normalsize
\newpage
\appendix

\noindent\textbf{Appendix organization.} We first provide the theoretical analysis and additional related work, then describe dataset preparation, followed by network and experimental details for reproducibility. We further report additional experiments, qualitative visualizations, and limitations.

\section{Theoretical Analysis}
\label{theory_analysis}
This section provides two formal observations that match the design logic of ControlFlow3D. The first explains why multi-view information is useful as a training-time control signal for the latent velocity field. The second explains why aligning image-side and 3D-side prototypes helps keep the visual controller geometrically consistent. These results are population-level statements under standard square-integrability and Lipschitz assumptions. They do not claim an optimization guarantee for the finite neural network. Consistent with the main method, multi-view variables are used only by the training-time controller, while inference remains point-only.

\subsection{Training-Time Multi-View Conditioning Reduces Oracle Flow-Matching Risk}
\label{theory_multiview}
\noindent\textbf{Assumption A.1.}
Let $X_0,X_1\in\mathbb{R}^{d}$ denote the sparse-aligned source state and dense-target latent state on the same sparse-query support, and let $U=X_1-X_0$ be the flow-matching target. For a sampled time $t$, let $X_t=(1-t)X_0+tX_1$. Let $c$ denote the sparse geometric condition and define $Z=(X_t,t,c)$ as the information available to the sparse-conditioned velocity network. Let $\mathcal{I}$ denote the multi-view image feature available during training. We assume $\mathbb{E}\|U\|_2^2<\infty$ and consider oracle predictors over all square-integrable measurable functions.

\noindent\textbf{Theorem A.1 (Value of training-time multi-view information).}
Under Assumption A.1, define the oracle sparse-conditioned and training-time multi-view-conditioned flow-matching risks as
\begin{align}
    \mathcal{R}^{*}_{base}
    &=
    \inf_{f}
    \mathbb{E}
    \left[
    \|U-f(Z)\|_2^2
    \right],\\
    \mathcal{R}^{*}_{ctrl}
    &=
    \inf_{g}
    \mathbb{E}
    \left[
    \|U-g(Z,\mathcal{I})\|_2^2
    \right].
\end{align}
Then
\begin{equation}
    \mathcal{R}^{*}_{base}
    -
    \mathcal{R}^{*}_{ctrl}
    =
    \mathbb{E}
    \left[
    \left\|
    \mathbb{E}[U\mid Z,\mathcal{I}]
    -
    \mathbb{E}[U\mid Z]
    \right\|_2^2
    \right]
    \ge 0.
    \label{eq:mv_risk_gap}
\end{equation}
Consequently, at the oracle level, conditioning on multi-view information cannot increase the optimal squared flow-matching risk. The inequality is strict exactly when
$\mathbb{E}[U\mid Z,\mathcal{I}]\ne\mathbb{E}[U\mid Z]$ with positive probability.

\noindent\textit{Proof.}
For any square-integrable random vector $U$ and conditioning variable $Y$, the minimizer of
$\mathbb{E}\|U-h(Y)\|_2^2$ over measurable $h$ is the conditional expectation
$h^*(Y)=\mathbb{E}[U\mid Y]$. This follows from the orthogonal projection property of conditional expectation in $L^2$. Applying this fact with $Y=Z$ and $Y=(Z,\mathcal{I})$ gives
\begin{align}
    \mathcal{R}^{*}_{base}
    &=
    \mathbb{E}
    \left[
    \left\|
    U-\mathbb{E}[U\mid Z]
    \right\|_2^2
    \right],\\
    \mathcal{R}^{*}_{ctrl}
    &=
    \mathbb{E}
    \left[
    \left\|
    U-\mathbb{E}[U\mid Z,\mathcal{I}]
    \right\|_2^2
    \right].
\end{align}
By the tower property,
$\mathbb{E}[\mathbb{E}[U\mid Z,\mathcal{I}]\mid Z]=\mathbb{E}[U\mid Z]$.
Define $\mu_Z=\mathbb{E}[U\mid Z]$ and $\mu_{Z,\mathcal{I}}=\mathbb{E}[U\mid Z,\mathcal{I}]$. Then
$U-\mathbb{E}[U\mid Z]$ decomposes into
\begin{equation}
    U-\mu_Z
    =
    \left(U-\mu_{Z,\mathcal{I}}\right)
    +
    \left(\mu_{Z,\mathcal{I}}-\mu_Z\right).
\end{equation}
The cross term is zero because
\begin{equation}
    \mathbb{E}
    \left[
    \left\langle
    U-\mu_{Z,\mathcal{I}},
    \mu_{Z,\mathcal{I}}-\mu_Z
    \right\rangle
    \right]
    =
    \mathbb{E}
    \left[
    \left\langle
    \mathbb{E}[U-\mu_{Z,\mathcal{I}}\mid Z,\mathcal{I}],
    \mu_{Z,\mathcal{I}}-\mu_Z
    \right\rangle
    \right]
    =
    0 .
\end{equation}
Taking squared norms and expectations therefore yields Eq.~\eqref{eq:mv_risk_gap}. Non-negativity follows from the squared norm, and strictness holds precisely when the second term is nonzero with positive probability. \hfill$\square$

\noindent\textbf{Implication.}
The theorem formalizes a population-level reason for using multi-view cues to control the latent velocity field during training. If the images contain information about the dense latent displacement that is not recoverable from the sparse point condition alone, a training-time multi-view controller can provide a lower-risk velocity signal than a sparse-only oracle. In ControlFlow3D, this does not introduce image inputs at inference: the multi-view signal is used to guide training, and the model predicts at inference from the sparse point cloud and sparse 3D features.

\subsection{Prototype Consistency Controls Velocity-Modulation Mismatch}
\label{theory_proto}
\noindent\textbf{Assumption A.2.}
Let $P^{img},P^{3D}\in\mathbb{R}^{K_p\times d_p}$ be image-side and 3D-side prototypes paired by the shared soft-KMeans centers, with each row normalized to unit Euclidean norm. Let
\begin{equation}
    \mathcal{L}_{\mathrm{proto}}
    =
    1-
    \frac{1}{K_p}
    \sum_{i=1}^{K_p}
    \left\langle
    P^{img}_{i},P^{3D}_{i}
    \right\rangle .
    \label{eq:proto_loss_theory}
\end{equation}
Let $m(P)=(\gamma(P),\beta(P))$ be the prototype-to-modulation map that produces affine velocity parameters, and assume it is $L$-Lipschitz:
\begin{equation}
    \left(
    \|\gamma(P)-\gamma(P')\|_2^2
    +
    \|\beta(P)-\beta(P')\|_2^2
    \right)^{1/2}
    \le
    L\|P-P'\|_F .
    \label{eq:lipschitz_modulation}
\end{equation}
For a base velocity $v$, define the affine controller
\begin{equation}
    \mathcal{A}_{m(P)}(v)=\gamma(P)\odot v+\beta(P).
\end{equation}

\noindent\textbf{Theorem A.2 (Prototype consistency bound).}
Under Assumption A.2, for any base velocity satisfying $\|v\|_{\infty}\le B$,
\begin{equation}
    \left\|
    \mathcal{A}_{m(P^{img})}(v)
    -
    \mathcal{A}_{m(P^{3D})}(v)
    \right\|_2
    \le
    L
    \sqrt{
    2K_p(B^2+1)\mathcal{L}_{\mathrm{proto}}
    } .
    \label{eq:proto_bound}
\end{equation}

\noindent\textit{Proof.}
Let
$\Delta_{\gamma}=\gamma(P^{img})-\gamma(P^{3D})$ and
$\Delta_{\beta}=\beta(P^{img})-\beta(P^{3D})$.
The controller difference is
\begin{equation}
    \mathcal{A}_{m(P^{img})}(v)-\mathcal{A}_{m(P^{3D})}(v)
    =
    \Delta_{\gamma}\odot v+\Delta_{\beta}.
\end{equation}
Using $\|v\|_{\infty}\le B$ and the Cauchy--Schwarz inequality,
\begin{align}
    \|\Delta_{\gamma}\odot v+\Delta_{\beta}\|_2
    &\le
    B\|\Delta_{\gamma}\|_2+\|\Delta_{\beta}\|_2 \\
    &\le
    \sqrt{B^2+1}
    \left(
    \|\Delta_{\gamma}\|_2^2+\|\Delta_{\beta}\|_2^2
    \right)^{1/2}.
    \label{eq:controller_difference}
\end{align}
By the Lipschitz condition in Eq.~\eqref{eq:lipschitz_modulation},
\begin{equation}
    \left(
    \|\Delta_{\gamma}\|_2^2+\|\Delta_{\beta}\|_2^2
    \right)^{1/2}
    \le
    L\|P^{img}-P^{3D}\|_F .
\end{equation}
Since each prototype is unit-normalized,
\begin{align}
    \|P^{img}-P^{3D}\|_F^2
    &=
    \sum_{i=1}^{K_p}
    \|P^{img}_{i}-P^{3D}_{i}\|_2^2 \\
    &=
    \sum_{i=1}^{K_p}
    \left(
    2-2
    \left\langle
    P^{img}_{i},P^{3D}_{i}
    \right\rangle
    \right)\\
    &=
    2K_p\mathcal{L}_{\mathrm{proto}}.
\end{align}
Substituting this identity into Eq.~\eqref{eq:controller_difference} proves Eq.~\eqref{eq:proto_bound}. \hfill$\square$

\noindent\textbf{Implication.}
The bound should be read as a smoothness statement for the prototype controller. If the same locally Lipschitz modulation map is evaluated on paired image-side and 3D-side prototypes, reducing their cosine mismatch also reduces the possible mismatch between the corresponding affine velocity modulations. This supports using $\mathcal{L}_{\mathrm{proto}}$ as a stability regularizer that discourages image-driven control from becoming inconsistent with geometry-supported 3D prototypes.

\section{Additional Related Work on Knowledge Distillation}
\label{appendix_kd_related_work}

\textbf{Knowledge distillation for point cloud learning.}
Knowledge distillation transfers information from richer predictions, features, or auxiliary modalities into a target model~\cite{hinton2015distilling}. In point cloud learning, prior methods adapt the transferred signal to 3D structure through point-level or voxel-level supervision~\cite{hou2022point,zhang2023pointdistiller}. Other works introduce local-structure, global-context, or cross-modal constraints to improve the student representation~\cite{li2023bidirectional,yan2022let,zhang2023pointmcd}. These methods mainly rely on direct imitation or auxiliary losses, such as matching logits, responses, attention maps, intermediate features, or visual supervisory signals.

\textbf{Distillation for point cloud reconstruction and upsampling.}
For reconstruction and upsampling, existing work transfers completion priors or vector-quantized codebook knowledge to improve dense geometry prediction~\cite{fan2022reconstruction,zhao2023cpu}. MVP-Net is closely related because it uses rendered multi-view depth images for upsampling~\cite{chen2024mvp}. However, it still requires inference-time depth images, extra preprocessing, and direct point-depth feature fusion. These methods show the value of geometric distillation, while remaining different from our training-time image distillation setting.

ControlFlow3D uses \emph{shared-prototype distillation} instead of direct feature imitation or direct concatenation of 2D and 3D features. During training, VGGT image tokens and Utonia dense-target tokens are projected into a common prototype space, where image-side and 3D-side prototypes are aligned without explicit pixel-to-point correspondence. The aligned prototypes modulate the latent velocity field during training, and inference uses only the sparse point cloud and sparse 3D features.

\section{Dataset Preparation}

Existing point cloud upsampling benchmarks, such as PU-GAN~\cite{li2019pu} and PU1K~\cite{qian2021pu}, are mainly designed for patch-based learning. Their source models are relatively limited, and each object is usually decomposed into many overlapping local patches. This design has two limitations. First, the network mainly observes isolated local neighborhoods rather than complete object-level context, making it difficult to model global structural consistency across different regions. Second, during inference, multiple upsampled patches are usually merged and then downsampled by FPS. Although this strategy reduces duplicated candidates, it may alter the original density distribution of the object. Since different regions of a shape do not necessarily have the same sampling density, fusion based on overlapping local patches may affect the preservation of fine geometric details.

To address these issues, we construct a new multi-modal point cloud upsampling dataset, denoted as \textbf{ShapeNetPU}. Unlike previous patch-only benchmarks, ShapeNetPU provides both whole-object context and non-overlapping patch-level upsampling pairs, making it suitable for global-to-local upsampling pipelines. Starting from the R2N2 release~\cite{choy20163d}, we collect 43,783 samples, each with 24 rendered views and camera metadata. For each object, we select eight views with relatively diverse camera poses as the multi-view image input. We then match each sample to its corresponding OBJ mesh in ShapeNetCore.v1~\cite{shapenet2015}.

We generate the point-cloud supervision directly from the matched ShapeNetCore.v1 OBJ meshes. For each mesh, we first perform area-weighted random surface sampling to obtain 50K surface points. This dense random set is then downsampled by farthest point sampling (FPS) to 20K points, which is used as the intermediate surface pool. The 50K-to-20K procedure reduces the randomness of direct low-count sampling while keeping the final pool well spread over the mesh surface.

From the 20K surface pool, we construct paired sparse and dense point clouds with aligned non-overlapping patches. We first run KMeans on the 20K points to obtain eight object-level spatial patches. For each patch, we apply a two-stage FPS strategy. In the first stage, points are over-sampled within each KMeans patch, using an over-sampling factor $\alpha$ before the final quota is enforced. The local FPS is initialized from the point closest to the object centroid, which gives a deterministic initial point for each patch. In the second stage, candidates from all patches are merged and selected by a global FPS procedure with a strict per-patch quota. This global second pass prevents any single patch from dominating the sampling process and alleviates local density inconsistency caused by uneven raw mesh tessellation or random surface sampling.

The final sparse input contains 2048 points, with eight patches of 256 points each, while the dense target contains 8192 points, with the corresponding eight patches of 1024 points each. Points are written in patch order, and the patch index lists record the contiguous vertex ranges, so the model can recover the sparse-dense patch correspondence without re-clustering during training. A stable seed derived from the sample path makes the preprocessing reproducible.

After preprocessing and filtering invalid cases, ShapeNetPU contains 35,762 valid 3D models from 13 ShapeNet categories. Each sample contains the whole-object sparse point cloud, dense target point cloud, multi-view images, and original mesh. The sparse and dense point clouds retain the standard $4\times$ patch upsampling setting with 256 input points and 1024 target points per local patch.

\begin{figure}[!htbp]
\centering
\includegraphics[width=1\linewidth]{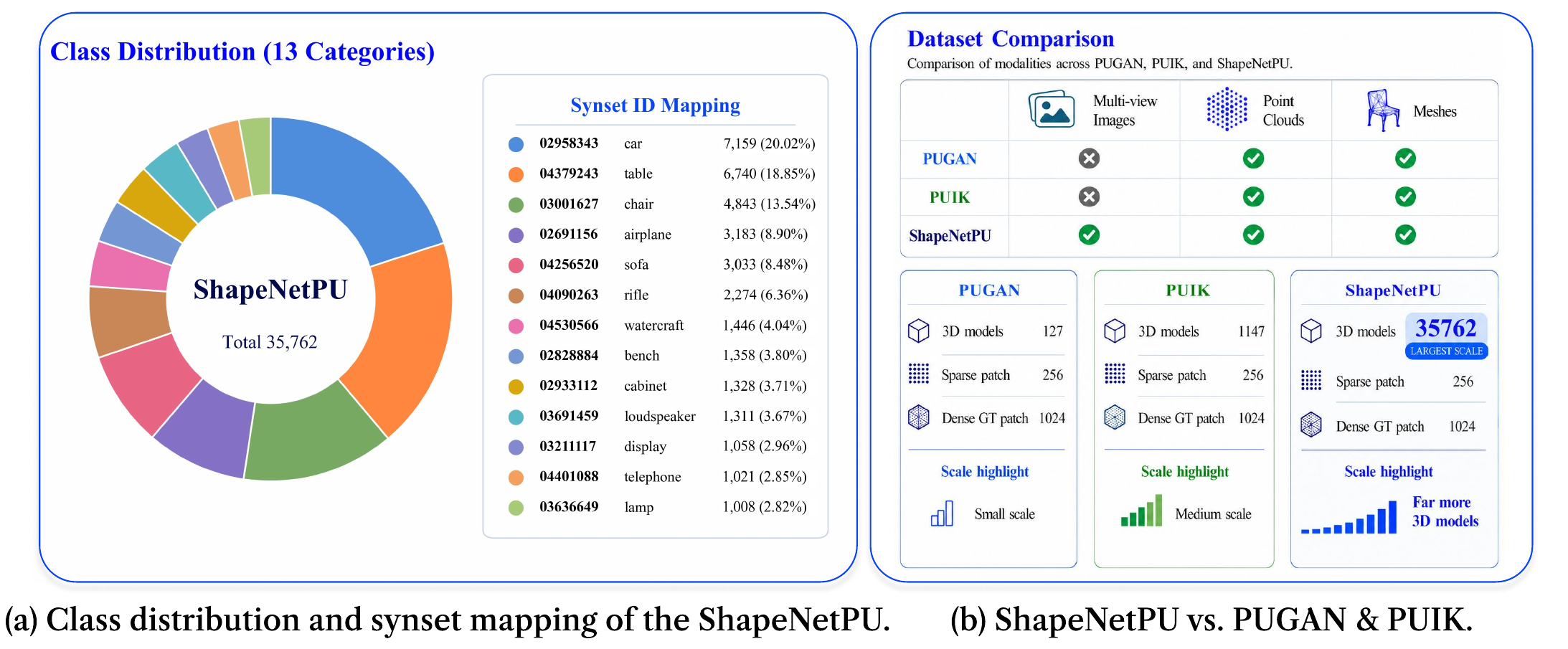}
\caption{Dataset statistics and benchmark comparison for ShapeNetPU. (a) The category distribution and ShapeNet synset mapping cover 13 object classes. (b) Compared with PU-GAN and PU1K, ShapeNetPU retains the standard sparse and dense point-cloud patch sizes while additionally providing multi-view images.}
\label{fig:dataset_distribution}
\end{figure}

\section{Network and Experimental Details}
\label{details}
\subsection{Utonia Details}
Utonia is used as the frozen 3D encoder that defines the structural latent space in our framework. Rather than asking the flow model to generate Euclidean coordinates directly, we first encode both sparse and dense point sets into a compact geometry-aware feature space. In implementation, the pretrained Utonia encoder processes point clouds with their original 3D coordinates and returns point-wise features, which are then projected to a 128-dimensional latent representation. These features are stored offline for both sparse inputs and dense ground truth point sets, so the training loop can focus on learning the transport dynamics rather than repeatedly running the frozen 3D foundation model. Keeping the encoder frozen preserves the geometric priors learned from large-scale pretraining and provides a stable target distribution for flow matching. If the encoder were jointly optimized with the generator, the target endpoint would keep changing during training and make the latent transport problem less well defined.

\begin{figure}[!htbp]
\centering
\includegraphics[width=0.95\linewidth]{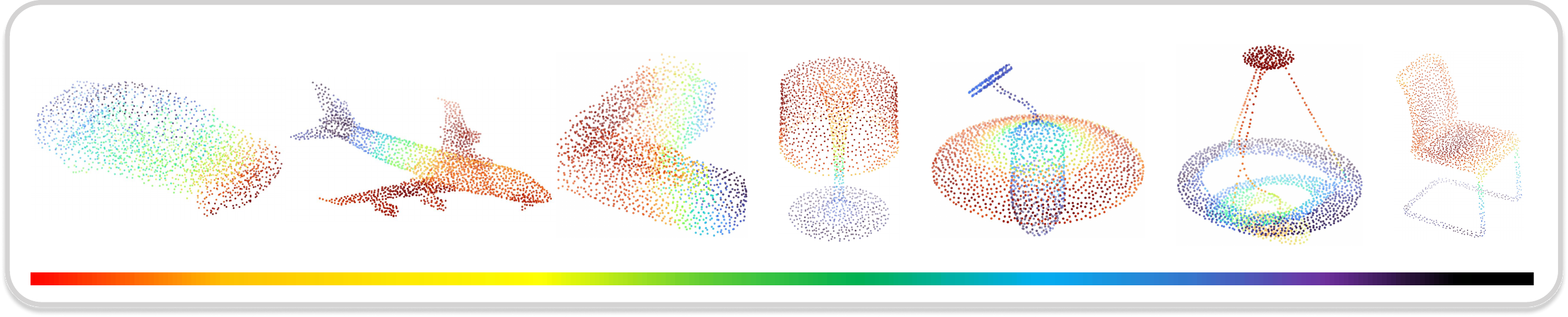}
\caption{Visualization of point-wise Utonia features on representative shapes. The feature responses vary smoothly over coherent geometric regions, indicating that the frozen 3D encoder provides a structured latent space for sparse-to-dense transport.}
\label{fig:utonia_feature_vis}
\end{figure}

The two Utonia feature streams play different roles. The sparse-aligned feature initializes the source state $x_0$, while the dense ground-truth feature defines the target state $x_1$. We align the dense feature to the sparse query support through three-nearest-neighbor interpolation, so each flow token has a dense-feature endpoint at an inference-available query location.

\subsection{Sparse Geometry Condition Network Details}
The sparse geometry condition network is deliberately lightweight and input-driven. We use a PointNet++ MSG encoder-decoder to extract both local and global descriptors from the sparse input point cloud. The set abstraction hierarchy samples $1024 \rightarrow 256 \rightarrow 64 \rightarrow 16$ points and uses two radii at each level to capture local neighborhoods of different sizes. Feature propagation then lifts the coarse descriptors back to the original sparse resolution, producing a point-wise local feature map with 256 channels. A max-pooled global descriptor is projected to the same 256-dimensional channel space.

From the sparse coordinates, we select 256 flow-token locations by FPS. The local condition is gathered or interpolated to these token locations, while the global descriptor is broadcast to every token. The final condition token for each query therefore contains three pieces of information:
\begin{equation}
    c_i =
    \operatorname{Concat}
    \left[
    q_i,
    f^{\mathrm{local}}_i,
    f^{\mathrm{global}}
    \right],
    \qquad
    c_i \in \mathbb{R}^{3+256+256}.
\end{equation}
The coordinate term specifies the token location in 3D space, the local descriptor describes the neighborhood around that token, and the global descriptor provides object-level context for the local patch. This conditioning design keeps each latent update tied to an explicit sparse geometry support, even though the flow evolves in feature space.

\subsection{Latent Flow Matching Network Details}
The latent velocity network predicts the displacement from the sparse latent state to the dense latent target. Given $x_t \in \mathbb{R}^{128 \times 256}$, the network first injects the time embedding through a FiLM-style conditional linear layer. We use the compact time code $[t,\sin(t),\cos(t)]$, which gives the velocity field both monotonic progress information and periodic basis functions for smoother conditioning. Four cross-attention blocks then repeatedly exchange information between latent flow tokens and sparse geometry condition tokens. Each block contains a multi-head cross-attention layer, a feed-forward layer, residual connections, layer normalization, and time-dependent affine modulation.

The prototype-guided controller is applied after the base velocity field during training. Multi-view image features are provided as patch tokens from VGGT. To avoid noisy token-wise 2D and 3D fusion, we compress the visual evidence into $K_{proto}=8$ shared semantic prototypes by default. These prototypes summarize local visual evidence and provide a compact basis for query-wise semantic modulation. Each soft-KMeans update uses cosine-normalized tokens and a temperature-controlled assignment, with four update iterations by default.

At training time, the prototype clustering observes both VGGT image tokens and query-aligned dense 3D target tokens. The image tokens provide cross-view cues for velocity modulation, while the dense target tokens constrain the shared prototypes using the geometry of the target surface. The soft query-to-prototype assignment distributes prototype-level scale and bias terms back to each query token, yielding the controlled training velocity
\begin{equation}
    \widetilde{v}_\theta
    =
    g(\mathcal{I},x_t,t)
    \odot
    v_\theta(x_t,t,c)
    +
    b(\mathcal{I},x_t,t).
\end{equation}
In the implementation, the prototype scale is parameterized as $1+0.35\tanh(\cdot)$ and the bias as $0.35\tanh(\cdot)$, so the control signal stays centered around the base velocity and remains bounded. We further regularize the image prototypes and 3D target prototypes with a cosine consistency loss, which discourages the visual controller from overemphasizing appearance cues that are not geometrically supported.

For inference, we integrate the learned ODE with a fixed 5-step Euler solver. Starting from the sparse-aligned latent state, the model repeatedly evaluates the learned velocity field and updates the latent tokens. A small residual endpoint MLP is then applied to the final state before decoding, which corrects endpoint bias accumulated by the finite-step rollout.

\subsection{Decoder Network Details}
The geometry decoder converts terminal latent tokens into dense point coordinates through a direct feed-forward reconstruction path. Given the sparse point cloud, sparse local condition, and predicted latent endpoint, it produces coordinates without adding a separate sampling or diffusion process. The latent flow outputs 256 global query tokens. For each sparse patch, these tokens are used as a flow-condition token set and fused with the patch's local sparse features inside the decoder.

We use a two-stage Mamba-based decoder with upsampling factors $(2,2)$. The first stage takes a sparse patch and produces a $2\times$ point set. It fuses the sparse local condition, positional embedding, and latent flow condition through a cross-attention layer, followed by a multi-scan Mamba block. The scan orders are Hilbert and reversed Hilbert curves, which preserve local neighborhoods while exposing the sequence model to two traversal directions. The second stage takes the $2\times$ intermediate points and predicts another $2\times$ point set using a pure Mamba block with $z$-order and reversed $z$-order scans. Finally, the stage-1 and stage-2 points are concatenated and passed through a position-aware residual head:
\begin{equation}
    \widehat{P}_{out}
    =
    \widehat{P}_{cand}
    +
    \tanh
    \left(
    \mathrm{MLP}
    \left[
    H_{cand},
    \mathrm{PE}(\widehat{P}_{cand})
    \right]
    \right).
\end{equation}
This design avoids treating upsampling as a one-shot coordinate expansion. The first stage predicts an intermediate denser point set, the second stage adds complementary points, and the residual head performs a final surface-aware adjustment. Because Mamba scales linearly with sequence length, the decoder can model longer surface dependencies than purely local MLP operators without incurring the quadratic cost of full self-attention.

Training and inference use the same patch organization defined by the dataset construction. The sparse input is divided into eight non-overlapping object patches. Each local patch is decoded with shared decoder weights, and the patch outputs are merged to form the final dense point cloud.

\subsection{Implementation Details}
We implement ControlFlow3D in PyTorch and train it with distributed data parallelism on four NVIDIA RTX 4090 GPUs. Unless otherwise specified, all models are trained for 500 epochs with a global batch size of 128, corresponding to 32 samples per GPU. The global test batch size is also set to 128. We set \texttt{OMP\_NUM\_THREADS}=1 and \texttt{MKL\_NUM\_THREADS}=1, and use four dataloader workers per training process with a prefetch factor of 2. We use AdamW as the optimizer with an initial learning rate of $2\times10^{-4}$ and weight decay $1\times10^{-4}$. The learning rate follows a cosine annealing schedule and decays to $1\times10^{-6}$ by the end of training. Gradients are clipped with a maximum norm of 1.0 for stable optimization.

\subsection{Training Protocol}
\label{appendix_training_protocol}
For ShapeNetPU training, each sample contains a sparse input of 2048 points and a dense target of 8192 points. Point coordinates are normalized by the sparse input centroid and scale before being processed by the model. We use a patch-based training protocol by default, where the sparse input is divided into eight non-overlapping patches of 256 points and the dense target is divided into the corresponding eight patches of 1024 points. The latent flow still observes the complete sparse object context, while the Mamba-based decoder receives patch-level supervision for local reconstruction.

The training objective combines the global coordinate reconstruction loss, local patch-level reconstruction loss, latent flow matching loss, and prototype consistency loss. We use the weighted objective
\begin{equation}
    \mathcal{L}
    =
    \mathcal{L}_{\mathrm{geo}}
    +
    \lambda_{\mathrm{FM}}\mathcal{L}_{\mathrm{FM}}
    +
    \lambda_{\mathrm{proto}}\mathcal{L}_{\mathrm{proto}},
\end{equation}
where $\mathcal{L}_{\mathrm{geo}}$ contains the global coordinate reconstruction term and the local patch-level reconstruction term. In the default setting, the local patch-level term is weighted by 0.35 within $\mathcal{L}_{\mathrm{geo}}$. The flow-matching loss weight is linearly scheduled from 0.003 to 0.0015, with a 20-epoch warm-up at the initial value. The prototype consistency loss weight is linearly scheduled from 0.01 to 0.005. During training, multi-view image features guide the latent velocity field.

\subsection{Loss-Weight Selection and Ablation}
\label{appendix_loss_weight_ablation}
The auxiliary latent losses have different numerical scales from the coordinate reconstruction loss. We therefore use small decaying weights: $\lambda_{\mathrm{FM}}$ provides latent endpoint supervision early in training, and $\lambda_{\mathrm{proto}}$ aligns image-side and 3D-side prototypes before becoming a lightweight regularizer. This ablation changes only these auxiliary weights and keeps the multi-view controller architecture fixed, so it differs from the ``w/o distillation'' setting in the main paper.

\begin{table}[!htbp]
\centering
\caption{Loss-weight ablation on ShapeNetPU under $4\times$ upsampling. Lower is better. The default setting is used in all main experiments.}
\label{tab:loss_weight_ablation}
\footnotesize
\setlength{\tabcolsep}{4.5pt}
\begin{tabular}{@{}lccccc@{}}
\toprule
Variant & $\lambda_{\mathrm{FM}}$ & $\lambda_{\mathrm{proto}}$ & CD$\downarrow$ & HD$\downarrow$ & P2F$\downarrow$ \\
\midrule
No auxiliary latent loss & 0 & 0 & 0.338 & 2.901 & 1.168 \\
Unit auxiliary weights & 1 & 1 & 0.347 & 3.018 & 1.241 \\
Weak FM & 0.0015\,\(\rightarrow\)\,0.00075 & 0.01\,\(\rightarrow\)\,0.005 & 0.292 & 2.388 & 0.891 \\
Strong FM & 0.006\,\(\rightarrow\)\,0.003 & 0.01\,\(\rightarrow\)\,0.005 & 0.290 & 2.371 & 0.878 \\
Weak prototype alignment & 0.003\,\(\rightarrow\)\,0.0015 & 0.005\,\(\rightarrow\)\,0.0025 & 0.287 & 2.342 & 0.860 \\
Strong prototype alignment & 0.003\,\(\rightarrow\)\,0.0015 & 0.02\,\(\rightarrow\)\,0.01 & 0.286 & 2.336 & 0.852 \\
\textbf{Default} & \textbf{0.003\,\(\rightarrow\)\,0.0015} & \textbf{0.01\,\(\rightarrow\)\,0.005} & \textbf{0.280} & \textbf{2.259} & \textbf{0.818} \\
\bottomrule
\end{tabular}
\end{table}

The default schedule gives the lowest errors in this ablation. Too little auxiliary supervision weakens latent alignment, while larger weights can overemphasize feature-space objectives relative to coordinate reconstruction.

\subsection{Additional Ablation Results}
\label{appendix_additional_ablation}
We further report controlled ablations for design choices and hyperparameters that are not exhaustively covered in the main paper. We evaluate all variants on the ShapeNetPU test set under $4\times$ upsampling. Unless a row explicitly changes a factor, we keep the frozen Utonia encoder, the weighted objective in Sec.~\ref{appendix_training_protocol}, the default $K_{proto}=8$, eight VGGT views, 256 sparse-query tokens, and a 5-step Euler solver.

\begin{table}[!htbp]
\centering
\caption{Additional ablations of the prototype controller on ShapeNetPU under $4\times$ upsampling. Lower is better.}
\label{tab:additional_controller_ablation}
\footnotesize
\setlength{\tabcolsep}{4.5pt}
\begin{tabular}{@{}llccc@{}}
\toprule
Variant & Main changed factor & CD$\downarrow$ & HD$\downarrow$ & P2F$\downarrow$ \\
\midrule
No image-guided controller & Base velocity only & 0.329 & 2.842 & 1.139 \\
All-token image fusion & Raw VGGT cross-attn & 0.303 & 2.501 & 0.958 \\
Image-only prototypes & No target 3D tokens & 0.296 & 2.421 & 0.900 \\
Hard query assignment & Nearest prototype & 0.291 & 2.366 & 0.872 \\
Scale-only controller & Use $\mathbf{g}\odot v_\theta$ only & 0.293 & 2.394 & 0.886 \\
Bias-only controller & Use $v_\theta+\mathbf{b}$ only & 0.289 & 2.352 & 0.862 \\
No prototype consistency & Set $\lambda_{\mathrm{proto}}=0$ & 0.294 & 2.386 & 0.884 \\
\textbf{Default} & Shared prototypes + $\mathbf{g},\mathbf{b}$ & \textbf{0.280} & \textbf{2.259} & \textbf{0.818} \\
\bottomrule
\end{tabular}
\end{table}

Table~\ref{tab:additional_controller_ablation} supports the shared-prototype controller. Removing image guidance gives the largest error, while direct all-token fusion and image-only prototypes remain weaker than the default. The scale-only and bias-only rows indicate that $\mathbf{g}$ and $\mathbf{b}$ provide complementary controls.

\begin{table}[!htbp]
\centering
\caption{Additional ablations of latent-state construction and geometry decoding on ShapeNetPU under $4\times$ upsampling. Lower is better.}
\label{tab:additional_latent_decoder_ablation}
\footnotesize
\setlength{\tabcolsep}{4.5pt}
\begin{tabular}{@{}llccc@{}}
\toprule
Variant & Main changed factor & CD$\downarrow$ & HD$\downarrow$ & P2F$\downarrow$ \\
\midrule
Gaussian source & Gaussian $x_0$ & 0.318 & 2.721 & 1.061 \\
Learned source & Learned $x_0$ & 0.301 & 2.494 & 0.941 \\
Train Utonia & Jointly update encoder & 0.305 & 2.548 & 0.956 \\
Independent target & Unaligned target tokens & 0.307 & 2.566 & 0.968 \\
1-NN target alignment & Replace 3-NN with 1-NN & 0.292 & 2.373 & 0.875 \\
No source perturbation & Set $\sigma=0$ & 0.287 & 2.326 & 0.850 \\
No endpoint residual & Remove endpoint MLP & 0.286 & 2.315 & 0.844 \\
No local patch term & No local term in $\mathcal{L}_{\mathrm{geo}}$ & 0.295 & 2.407 & 0.909 \\
\textbf{Default} & Frozen Utonia + 3-NN alignment & \textbf{0.280} & \textbf{2.259} & \textbf{0.818} \\
\bottomrule
\end{tabular}
\end{table}

Table~\ref{tab:additional_latent_decoder_ablation} shows that the default latent construction is the most reliable among the tested variants. Gaussian or learned source states remove sparse Utonia initialization, joint Utonia training destabilizes the target latent features, and independent target support breaks the query alignment used at inference. Removing the local patch term mainly increases P2F, consistent with its role in local surface supervision.

\begin{table}[!htbp]
\centering
\caption{Additional budget ablations on ShapeNetPU under $4\times$ upsampling. Each block changes one factor while keeping all other defaults fixed. Lower is better.}
\label{tab:additional_budget_ablation}
\footnotesize
\setlength{\tabcolsep}{3pt}
\begin{tabular}{p{0.20\linewidth}p{0.30\linewidth}ccc}
\toprule
Factor & Setting & CD$\downarrow$ & HD$\downarrow$ & P2F$\downarrow$ \\
\midrule
\multicolumn{5}{c}{\textit{Number of training views}} \\
Views & 1 view & 0.306 & 2.536 & 0.994 \\
Views & 2 views & 0.297 & 2.430 & 0.918 \\
Views & 4 views & 0.287 & 2.328 & 0.850 \\
Views & 12 views & 0.284 & 2.302 & 0.838 \\
Views & \textbf{8 views (default)} & \textbf{0.280} & \textbf{2.259} & \textbf{0.818} \\
\midrule
\multicolumn{5}{c}{\textit{Number of sparse-query tokens}} \\
Query tokens & 128 & 0.295 & 2.402 & 0.893 \\
Query tokens & 384 & 0.284 & 2.296 & 0.839 \\
Query tokens & 512 & 0.288 & 2.337 & 0.858 \\
Query tokens & \textbf{256 (default)} & \textbf{0.280} & \textbf{2.259} & \textbf{0.818} \\
\midrule
\multicolumn{5}{c}{\textit{Euler steps for latent rollout}} \\
Euler steps & 1 step & 0.302 & 2.490 & 0.944 \\
Euler steps & 3 steps & 0.286 & 2.320 & 0.850 \\
Euler steps & 7 steps & 0.283 & 2.292 & 0.835 \\
Euler steps & 10 steps & 0.286 & 2.319 & 0.846 \\
Euler steps & \textbf{5 steps (default)} & \textbf{0.280} & \textbf{2.259} & \textbf{0.818} \\
\bottomrule
\end{tabular}
\end{table}

The budget ablations justify the default setting. Eight views, 256 query tokens, and a 5-step Euler solver give the lowest errors among the tested budgets, while larger settings add computation without consistent gains.

\subsection{Supervised Fine-Tuning on PU-GAN and PU1K}
To evaluate transferability on standard public benchmarks, we further perform supervised fine-tuning on PU-GAN~\cite{li2019pu} and PU1K~\cite{qian2021pu}. Since these benchmarks do not provide synchronized multi-view images, fine-tuning is conducted in a sparse-only setting starting from the supervised ShapeNetPU checkpoint. Both benchmarks are trained on the official $4\times$ patch training sets: each sparse training patch contains 256 points, and the corresponding dense target contains 1024 points. The official PU-GAN training set contains 24K patch pairs, and the official PU1K training set contains 69K patch pairs. Sparse Utonia features are used as the feature initialization for the sparse input points.

We fine-tune all model parameters with the AdamW optimizer. The base learning rate is $2\times10^{-6}$, with weight decay $1\times10^{-5}$ and betas $(0.9,0.95)$. Following the different roles of model components, we use learning-rate multipliers of 0.5, 1.0, and 1.5 for the sparse geometry backbone, latent flow network, and Mamba decoder, respectively. The learning rate follows cosine annealing with a minimum value of $0.1$ times the base learning rate. Gradients are clipped with maximum norm 1.0. We use two RTX 4090 GPUs for each fine-tuning run, with a global batch size of 128. PU-GAN is fine-tuned for 50 epochs, while PU1K is fine-tuned for 40 epochs.

The supervised fine-tuning objective is the Chamfer-distance reconstruction loss computed after a 5-step sparse-only rollout. During testing, we follow the patch-based evaluation protocol: each 2048-point input shape is divided into 256-point patches, patch predictions are merged, and FPS is used to match the target output size of 8192 points. Testing is performed every epoch with patch size 256 and patch rate 3.0. This protocol keeps the fine-tuning and public-benchmark evaluation aligned with the patch-based setting of PU-GAN and PU1K.

\subsection{Inference and Evaluation}
At inference time, ControlFlow3D takes only the sparse point cloud as the raw input, and sparse 3D features are computed from this point input. We use a fixed 5-step Euler solver for latent flow integration. The predicted terminal latent tokens are then decoded by the Mamba-based geometry decoder to produce the dense point cloud. For the default $4\times$ setting on ShapeNetPU, the model upsamples 2048 input points to 8192 output points. For larger ratios, we follow the same model rollout strategy used in the main experiments.

Following common point cloud upsampling protocols, we report Chamfer Distance (CD), Hausdorff Distance (HD), and point-to-surface distance (P2F). CD and HD are computed between the predicted and ground-truth point sets in the normalized coordinate frame. For P2F, predictions are transformed back to the original object coordinate frame and evaluated against the retained mesh surface. Lower values indicate better performance for all metrics.

\section{Additional Experiments and Visualizations}
This section provides supplementary evidence beyond the main quantitative tables. We include additional visual comparisons, robustness tests under synthetic input perturbations, and arbitrary-rate evaluations. These experiments examine whether the learned sparse-to-dense mapping remains stable when the input distribution or target upsampling ratio changes, while keeping the inference setting point-only.

\subsection{Additional Qualitative Results}
We provide additional qualitative examples on ShapeNetPU and PU-GAN. The ShapeNetPU examples cover object-level cases with synchronized training views, while the PU-GAN examples evaluate the fine-tuned model under the standard patch-based public benchmark setting.

\begin{figure}[!htbp]
\centering
\includegraphics[width=1\linewidth]{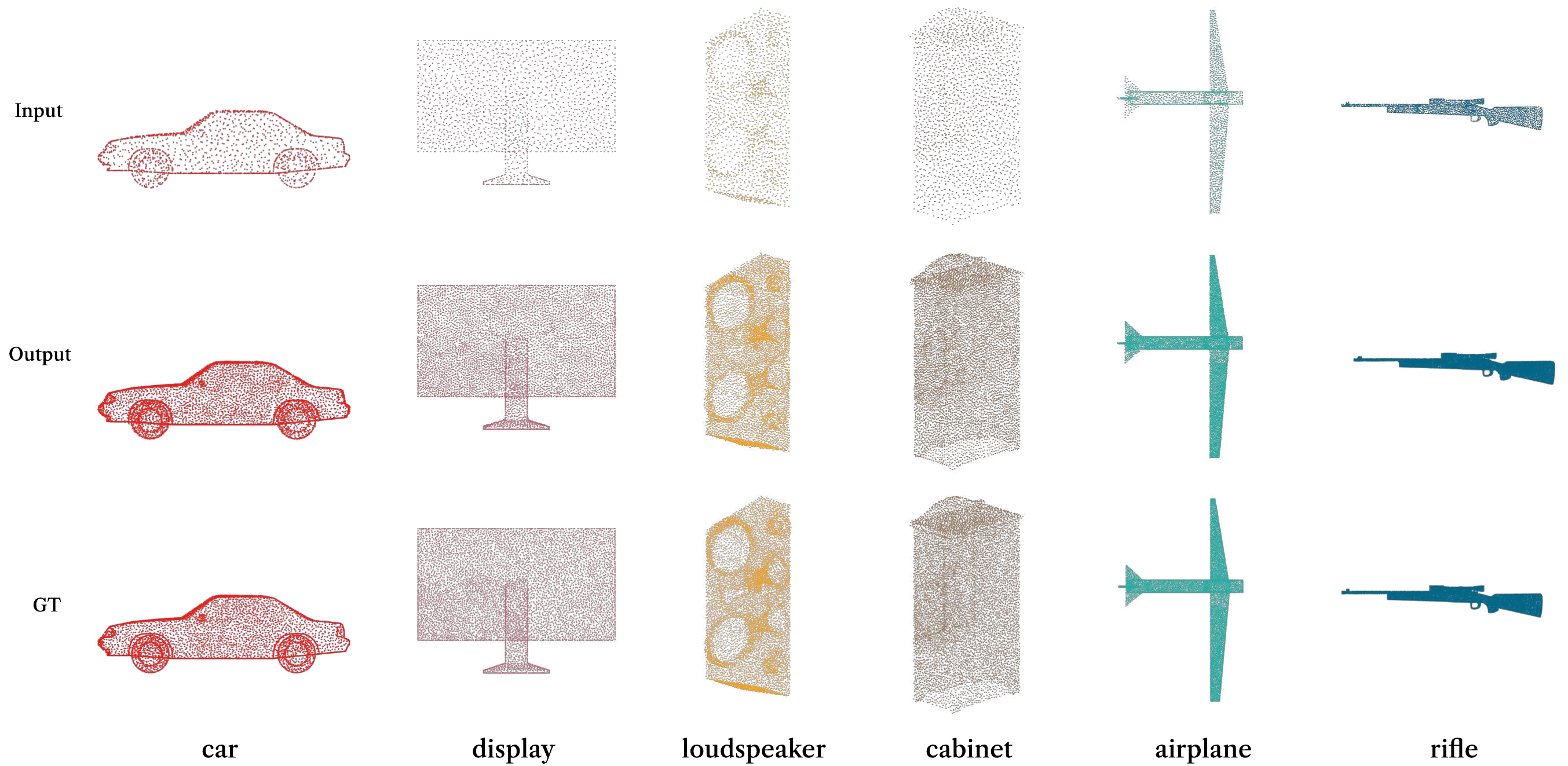}
\caption{More visualization results on ShapeNetPU.}
\label{fig:shapenetpu_more_categories}
\end{figure}

\begin{figure}[!htbp]
\centering
\includegraphics[width=1\linewidth]{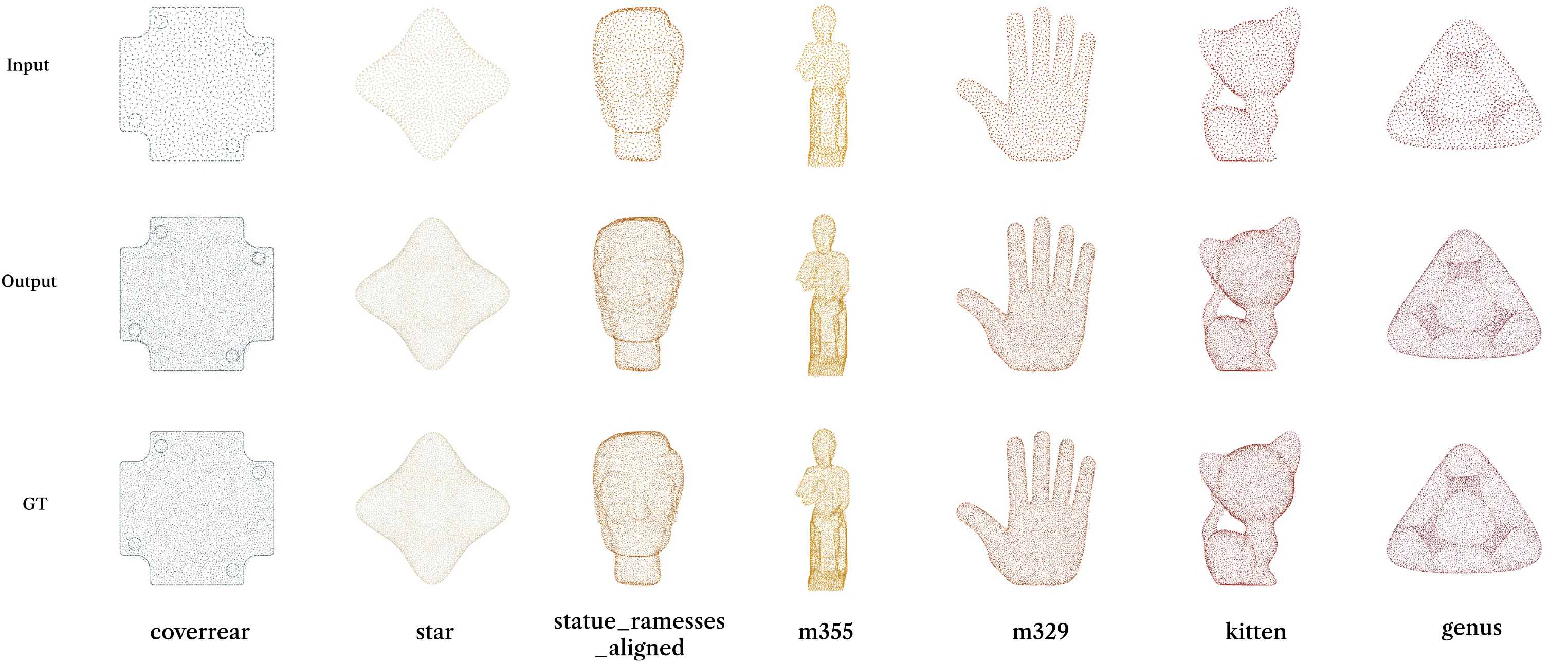}
\caption{More visualization results on PU-GAN under the $4\times$ upsampling setting.}
\label{fig:pugan_more_vis}
\end{figure}

Across these examples, ControlFlow3D generally preserves the global object structure while adding points along visible surfaces and thin components. The PU-GAN results further indicate that supervised fine-tuning transfers the learned decoder to regular patch-based surfaces without requiring image inputs at inference.

\subsection{Robustness to Random Noise}
Sparse inputs acquired in practical settings often contain measurement noise. To evaluate robustness, we add random Gaussian noise to ShapeNetPU inputs and perform $4\times$ upsampling under four noise levels, $\tau=0.01$, $\tau=0.02$, $\tau=0.05$, and $\tau=0.10$.

\begin{table}[!htbp]
\centering
\caption{Noise robustness comparison on ShapeNetPU under $4\times$ upsampling. Lower values are better.}
\label{tab:noise_robustness_4x}
\footnotesize
\setlength{\tabcolsep}{3pt}
\begin{tabular*}{\linewidth}{@{\extracolsep{\fill}}lcccccccccccc@{}}
\toprule
Noise Levels
& \multicolumn{3}{c}{$\tau=0.01$}
& \multicolumn{3}{c}{$\tau=0.02$}
& \multicolumn{3}{c}{$\tau=0.05$}
& \multicolumn{3}{c}{$\tau=0.10$} \\
\cmidrule(lr){2-4} \cmidrule(lr){5-7} \cmidrule(lr){8-10} \cmidrule(lr){11-13}
Methods
& CD$\downarrow$ & HD$\downarrow$ & P2F$\downarrow$
& CD$\downarrow$ & HD$\downarrow$ & P2F$\downarrow$
& CD$\downarrow$ & HD$\downarrow$ & P2F$\downarrow$
& CD$\downarrow$ & HD$\downarrow$ & P2F$\downarrow$ \\
\midrule
Grad-PU~\cite{he2023grad}  
& 0.638 & 5.462 & 4.684
& 1.176 & 8.936 & 9.752
& 1.384 & 10.284 & 15.876
& 1.692 & 11.438 & 18.420 \\

APU-LDI~\cite{li2024learning}  
& 0.586 & 5.108 & 3.986
& 1.084 & 8.412 & 8.926
& 1.276 & 9.672 & 14.286
& 1.548 & 10.864 & 17.126 \\

PUDM~\cite{qu2024conditional}     
& 0.542 & 4.946 & 4.214
& 0.986 & 7.846 & 8.684
& 1.168 & 9.142 & 13.768
& 1.406 & 10.326 & 16.204 \\

PUFM~\cite{liu2026pufm}     
& 0.518 & 4.812 & 4.386
& 0.948 & 7.632 & 8.984
& 1.126 & 9.284 & 14.026
& 1.358 & 10.052 & 15.862 \\

\midrule
Ours             
& \textbf{0.431} & \textbf{3.620} & \textbf{2.384}
& \textbf{0.792} & \textbf{6.684} & \textbf{5.426}
& \textbf{0.982} & \textbf{8.216} & \textbf{9.384}
& \textbf{1.138} & \textbf{9.428} & \textbf{11.562} \\
\bottomrule
\end{tabular*}
\end{table}

This experiment evaluates how performance changes when the sparse input is perturbed around the underlying surface. All methods degrade as the noise level increases, but ControlFlow3D keeps lower CD, HD, and P2F across the tested levels. This suggests that the structured sparse features and geometric decoder reduce sensitivity to small coordinate perturbations.

\subsection{Quantitative comparison of Arbitrary-Rate Upsampling}
In addition to the arbitrary-rate visual results shown in the main paper, we report quantitative results on ShapeNetPU. The same trained model is evaluated from $5\times$ to $20\times$ upsampling to test whether the learned generation process generalizes beyond the native $4\times$ setting.

\begin{table}[!htbp]
\centering
\caption{Arbitrary-rate upsampling comparison with PUFM and APU-LDI on ShapeNetPU. Lower values are better.}
\label{tab:arbitrary_rates_shapenetpu}
\footnotesize
\setlength{\tabcolsep}{4.5pt}
\begin{tabular*}{\textwidth}{@{\extracolsep{\fill}}lccccccccc}
\toprule
Rates
& \multicolumn{3}{c}{PUFM~\cite{liu2026pufm}}
& \multicolumn{3}{c}{APU-LDI~\cite{li2024learning}}
& \multicolumn{3}{c}{Ours} \\
\cmidrule(lr){2-4} \cmidrule(lr){5-7} \cmidrule(lr){8-10}
& CD$\downarrow$ & HD$\downarrow$ & P2F$\downarrow$
& CD$\downarrow$ & HD$\downarrow$ & P2F$\downarrow$
& CD$\downarrow$ & HD$\downarrow$ & P2F$\downarrow$ \\
\midrule
$5\times$  & 0.325 & 3.260 & 1.707 & 0.342 & 3.105 & 1.505 & \textbf{0.268} & \textbf{2.300} & \textbf{0.893} \\
$7\times$  & 0.306 & 3.238 & 1.762 & 0.311 & 3.025 & 1.620 & \textbf{0.258} & \textbf{2.410} & \textbf{0.970} \\
$8\times$  & 0.299 & 3.252 & 1.790 & 0.299 & 3.045 & 1.600 & \textbf{0.251} & \textbf{2.335} & \textbf{1.002} \\
$15\times$ & 0.268 & 3.226 & 1.890 & 0.251 & 2.970 & 1.745 & \textbf{0.225} & \textbf{2.465} & \textbf{1.150} \\
$20\times$ & 0.255 & 3.178 & 1.945 & 0.238 & 2.925 & 1.860 & \textbf{0.217} & \textbf{2.575} & \textbf{1.210} \\
\bottomrule
\end{tabular*}
\end{table}

ControlFlow3D obtains the lowest CD, HD, and P2F among the compared methods for all evaluated ratios. The gains are most pronounced in P2F, indicating that the generated points remain closer to the underlying surface when the requested output density increases.

\section{Discussion}
\label{appendix_limitations}
\subsection{Failure Analysis}
The sparse-to-dense mapping remains under-constrained when the input provides weak geometric evidence or when the relevant structure is only partially observed.

\begin{figure}[!htbp]
\centering
\includegraphics[width=1\linewidth]{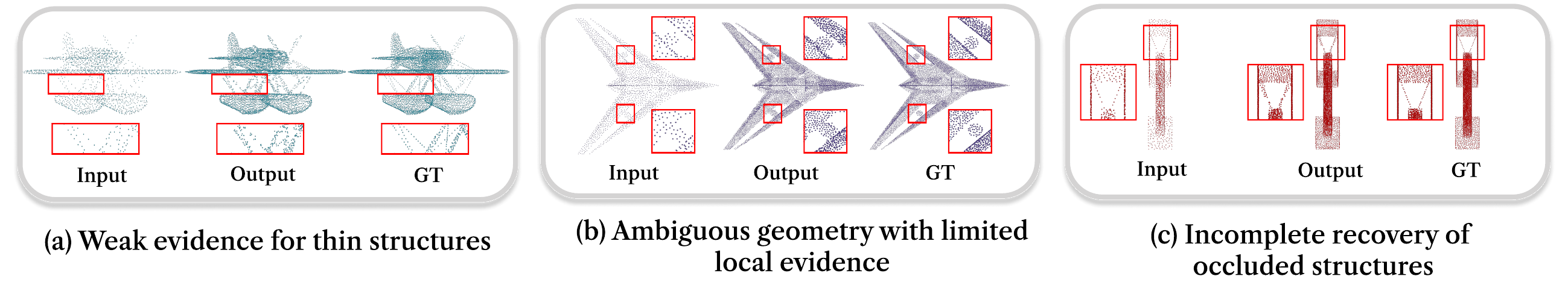}
\caption{Failure analysis of ControlFlow3D. Red boxes highlight challenging regions in the sparse input, our output, and the dense ground truth (GT).}
\label{fig:failure_analysis}
\end{figure}

The first pattern is weak evidence for thin structures, as shown in Figure~\ref{fig:failure_analysis}(a). Narrow parts such as airplane wings, chair legs, lamp supports, and similar elongated components may contain only a small number of sparse input points. The latent flow can infer the coarse structural identity, but the decoder may place points slightly off the true surface or recover an incomplete structure compared with the ground truth.

The second pattern is ambiguity caused by limited local evidence, as shown in Figure~\ref{fig:failure_analysis}(b). When several dense surfaces are compatible with the same sparse observations, the model may generate a geometrically plausible averaged solution. This behavior can blur sharp local boundaries or create small density leakage between adjacent semantic parts.

The third pattern is incomplete recovery of occluded or sparsely visible structures, as shown in Figure~\ref{fig:failure_analysis}(c). Multi-view distillation provides additional boundary and part-level cues during training, and these cues are encoded by the learned velocity field. At inference, the model does not require 2D inputs. It predicts from the sparse point cloud and the distilled prior. If a structure is poorly represented in the point input and not reliably encoded in the learned prior, the output may preserve the main visible region while missing part of the occluded or narrow component.

\subsection{Limitations}
Our framework relies on pretrained visual and 3D foundation models. This provides the latent flow with a structured representation space, but it also makes the method dependent on the coverage and biases of these encoders. If the frozen 3D encoder under-represents a particular geometry family, the corresponding latent targets may not faithfully preserve its fine structures. Similarly, if the rendered multi-view images do not reveal important regions because of occlusion or limited viewpoints, the prototype controller cannot recover information that is absent from the training signals.

Finally, the decoder is designed around a native $4\times$ decoding path. Larger arbitrary ratios are handled by repeated rollout or post-processing, which keeps the architecture simple but may increase inference cost and affect density preservation at very high ratios. A future direction is to condition the latent flow and decoder directly on the target density ratio.

\section{Broader Impact}
\label{appendix_broader_impact}
ControlFlow3D leverages multi-view and 3D foundation model knowledge during training while requiring only sparse point clouds at inference. This setting is relevant to applications where dense geometric observations are expensive or difficult to obtain, such as robotics, autonomous driving, augmented reality, digital twins, and cultural heritage digitization. Denser and more surface-consistent point clouds may benefit downstream perception, reconstruction, and visualization systems. However, point cloud upsampling also carries potential risks because the generated points are inferred from learned priors rather than directly measured by sensors. In safety-critical applications, such as autonomous navigation or industrial inspection, upsampled geometry should therefore be treated as model-assisted estimation rather than ground-truth observation. Although our method does not use identity-related data, real-world 3D scans may still contain sensitive spatial information, and responsible use should consider privacy, fairness, and uncertainty in practical use.

\newpage
\newpage
\section*{NeurIPS Paper Checklist}

\begin{enumerate}

\item {\bf Claims}
    \item[] Question: Do the main claims made in the abstract and introduction accurately reflect the paper's contributions and scope?
    \item[] Answer: \answerYes{} % Replace by \answerYes{}, \answerNo{}, or \answerNA{}.
    \item[] Justification: The main contributions and scope have been included in the abstract and introduction.

\item {\bf Limitations}
    \item[] Question: Does the paper discuss the limitations of the work performed by the authors?
    \item[] Answer: \answerYes{} % Replace by \answerYes{}, \answerNo{}, or \answerNA{}.
    \item[] Justification: The discussion on the limitations of our work is stated in Sec.~\ref{limitationandconclusion} and Appendix~\ref{appendix_limitations}.

\item {\bf Theory assumptions and proofs}
    \item[] Question: For each theoretical result, does the paper provide the full set of assumptions and a complete (and correct) proof?
    \item[] Answer: \answerYes{} % Replace by \answerYes{}, \answerNo{}, or \answerNA{}.
    \item[] Justification: We provide the assumptions, theorem statements, and complete proofs for the multi-view conditioning result and the prototype-consistency stability bound in Appendix~\ref{theory_analysis}.

    \item {\bf Experimental result reproducibility}
    \item[] Question: Does the paper fully disclose all the information needed to reproduce the main experimental results of the paper to the extent that it affects the main claims and/or conclusions of the paper (regardless of whether the code and data are provided or not)?
    \item[] Answer: \answerYes{} % Replace by \answerYes{}, \answerNo{}, or \answerNA{}.
    \item[] Justification: Comprehensive experimental details necessary for reproducibility, including datasets and model configurations, are thoroughly detailed in Sec.~\ref{exp}  and Appendix~\ref{details}. These details collectively ensure that the main experimental results supporting the claim and conclusions of the paper can be independently reproduced.

\item {\bf Open access to data and code}
    \item[] Question: Does the paper provide open access to the data and code, with sufficient instructions to faithfully reproduce the main experimental results, as described in supplemental material?
    \item[] Answer: \answerYes{} % Replace by \answerYes{}, \answerNo{}, or \answerNA{}.
    \item[] Justification: The code and dataset are publicly available via the link provided immediately after the abstract, along with instructions for reproducing the main experimental results.

\item {\bf Experimental setting/details}
    \item[] Question: Does the paper specify all the training and test details (e.g., data splits, hyperparameters, how they were chosen, type of optimizer) necessary to understand the results?
    \item[] Answer: \answerYes{} % Replace by \answerYes{}, \answerNo{}, or \answerNA{}.
    \item[] Justification: We specify the datasets, metrics, training protocol, optimizer, learning-rate schedule, batch size, model configurations, fine-tuning settings, and inference/evaluation details in Sec.~\ref{exp} and Appendix~\ref{details}.

\item {\bf Experiment statistical significance}
    \item[] Question: Does the paper report error bars suitably and correctly defined or other appropriate information about the statistical significance of the experiments?
    \item[] Answer: \answerYes{} % Replace by \answerYes{}, \answerNo{}, or \answerNA{}.
    \item[] Justification: We report the performance in terms of chamfer distance (CD), hausdorff distance (HD) and the point-to-surface distance (P2F) in Sec.~\ref{exp}.

\item {\bf Experiments compute resources}
    \item[] Question: For each experiment, does the paper provide sufficient information on the computer resources (type of compute workers, memory, time of execution) needed to reproduce the experiments?
    \item[] Answer: \answerYes{} % Replace by \answerYes{}, \answerNo{}, or \answerNA{}.
    \item[] Justification: We primarily executed all experiments relying on GPU resources, with details of the specific GPUs used provided in Sec.~\ref{exp}, and corresponding execution times reported in Fig.~\ref{fig:Fig1}(b).
    
\item {\bf Code of ethics}
    \item[] Question: Does the research conducted in the paper conform, in every respect, with the NeurIPS Code of Ethics \url{https://neurips.cc/public/EthicsGuidelines}?
    \item[] Answer: \answerYes{} % Replace by \answerYes{}, \answerNo{}, or \answerNA{}.
    \item[] Justification: We have carefully reviewed the NeurIPS Code of Ethics, and we affirm that the research conducted in our paper adheres to the NeurIPS Code of Ethics in every respect.

\item {\bf Broader impacts}
    \item[] Question: Does the paper discuss both potential positive societal impacts and negative societal impacts of the work performed?
    \item[] Answer: \answerYes{} % Replace by \answerYes{}, \answerNo{}, or \answerNA{}.
    \item[] Justification: We present broader impacts of our work in Appendix~\ref{appendix_broader_impact}.
    
\item {\bf Safeguards}
    \item[] Question: Does the paper describe safeguards that have been put in place for responsible release of data or models that have a high risk for misuse (e.g., pre-trained language models, image generators, or scraped datasets)?
    \item[] Answer: \answerNA{} % Replace by \answerYes{}, \answerNo{}, or \answerNA{}.
    \item[] Justification: To our knowledge, our work is not at risk of this issue.

\item {\bf Licenses for existing assets}
    \item[] Question: Are the creators or original owners of assets (e.g., code, data, models), used in the paper, properly credited and are the license and terms of use explicitly mentioned and properly respected?
    \item[] Answer: \answerYes{} % Replace by \answerYes{}, \answerNo{}, or \answerNA{}.
    \item[] Justification: We properly cite related works, codes and datasets.

\item {\bf New assets}
    \item[] Question: Are new assets introduced in the paper well documented and is the documentation provided alongside the assets?
    \item[] Answer: \answerYes{} % Replace by \answerYes{}, \answerNo{}, or \answerNA{}.
    \item[] Justification: ShapeNetPU is documented in Appendix C and the accompanying dataset card, which describes the dataset structure, splits, file formats, download procedure, and terms of use.

\item {\bf Crowdsourcing and research with human subjects}
    \item[] Question: For crowdsourcing experiments and research with human subjects, does the paper include the full text of instructions given to participants and screenshots, if applicable, as well as details about compensation (if any)? 
    \item[] Answer: \answerNA{} % Replace by \answerYes{}, \answerNo{}, or \answerNA{}.
    \item[] Justification: Our experiments focus on point cloud data and do not include crowdsourcing or involve research with human subjects.

\item {\bf Institutional review board (IRB) approvals or equivalent for research with human subjects}
    \item[] Question: Does the paper describe potential risks incurred by study participants, whether such risks were disclosed to the subjects, and whether Institutional Review Board (IRB) approvals (or an equivalent approval/review based on the requirements of your country or institution) were obtained?
    \item[] Answer: \answerNA{} % Replace by \answerYes{}, \answerNo{}, or \answerNA{}.
    \item[] Justification: The research presented in the paper does not involve human subjects or crowdsourcing experiments.

\item {\bf Declaration of LLM usage}
    \item[] Question: Does the paper describe the usage of LLMs if it is an important, original, or non-standard component of the core methods in this research? Note that if the LLM is used only for writing, editing, or formatting purposes and does \emph{not} impact the core methodology, scientific rigor, or originality of the research, declaration is not required.
    %this research? 
    \item[] Answer: \answerNA{} % Replace by \answerYes{}, \answerNo{}, or \answerNA{}.
    \item[] Justification: We do not involve LLMs as any important, original, or non-standard components.

\end{enumerate}

\end{document}